%% file: 0-main.tex
\documentclass[letterpaper, 10 pt, conference]{ieeeconf}  
\usepackage{booktabs}
\usepackage{multirow}
\usepackage{authblk}

\IEEEoverridecommandlockouts                              

\usepackage{amsmath} 
\usepackage{amssymb}  
\usepackage{xspace}  
\usepackage{algorithm}
\usepackage[noend]{algpseudocode}
\usepackage{graphicx}
\usepackage{xcolor}
\usepackage{caption}
\usepackage{url}
\usepackage{placeins}

\usepackage[noadjust]{cite}

\newcommand{\algo}{\mbox{IMPASTO}\xspace}
\title{\LARGE \bf
IMPASTO: Integrating Model-Based Planning with \\ Learned Dynamics Models for Robotic Oil Painting Reproduction
}
\author{
Yingke Wang$^{1}$, Hao Li$^{1}$, Yifeng Zhu$^{2}$, Hong-Xing Yu$^{1}$, \\
Ken Goldberg$^{3}$, Li Fei-Fei$^{1}$, Jiajun Wu$^{1}$, Yunzhu Li$^{4}$, and Ruohan Zhang$^{1}$\\
\thanks{$^{1}$Stanford University.}%
\thanks{$^{2}$The University of Texas at Austin.}%
\thanks{$^{3}$University of California, Berkeley.}%
\thanks{$^{4}$Columbia University.}%
}

\begin{document}

\makeatletter
\let\@oldmaketitle\@maketitle
\renewcommand{\@maketitle}{\@oldmaketitle
\begin{center}
    \centering
    \includegraphics[width=1\linewidth]{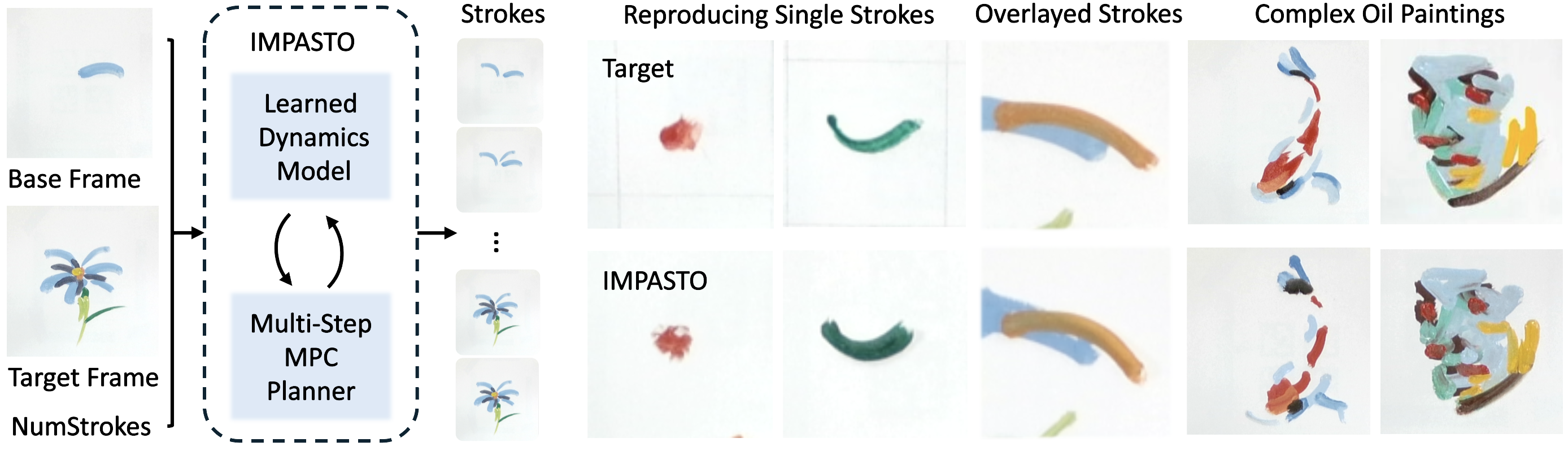}
    \captionof{figure}{\algo is a robotic oil painting system that integrates learned neural dynamics models with model-based planning algorithms to replicate human artists' brushstrokes and artworks. }
    \label{fig:pull}
    \vspace{-0.7cm}
\end{center}}
\makeatother

\maketitle
\addtocounter{figure}{-1}

\thispagestyle{empty}
\pagestyle{empty}


\begin{abstract}
Robotic reproduction of oil paintings using soft brushes and pigments requires force-sensitive control of deformable tools, prediction of brushstroke effects, and multi-step stroke planning, often without human step-by-step demonstrations or faithful simulators. Given only a sequence of target oil painting images, can a robot infer and execute the stroke trajectories, forces, and colors needed to reproduce it? We present \algo, a robotic oil-painting system that integrates learned pixel dynamics models with model-based planning. The dynamics models predict canvas updates from image observations and parameterized stroke actions; a receding-horizon model predictive control optimizer then plans trajectories and forces, while a force-sensitive controller executes strokes on a 7-DoF robot arm. \algo integrates low-level force control, learned dynamics models, and high-level closed-loop planning, learns solely from robot self-play, and approximates human artists' single-stroke datasets and multi-stroke artworks, outperforming baselines in reproduction accuracy. Project website: \url{https://impasto-robopainting.github.io/}.
\end{abstract}


\input{1-intro}
\input{2-related}

\input{3-method}
\input{4-results}
\input{5-conclude}

\section*{ACKNOWLEDGMENT}
This work is in part supported by the Stanford Institute for Human-Centered AI (HAI), the Schmidt Futures Senior Fellows grant, ONR MURI N00014-21-1-2801, ONR MURI N00014-22-1-2740, ONR MURI N00014-24-1-2748, and NSF RI \#2211258.

\bibliographystyle{IEEEtran}
\bibliography{ref,ref-dyn}

\clearpage
\appendix

\input{9-appendix}

\end{document}

%% file: 1-intro.tex
\section{Introduction}
Robot painting reproduction---enabling machines to accurately replicate physical artworks with brushes and pigments---has a rich history and presents unique technical challenges \cite{scalera2024history,karimov2023robot}. In modern times, these robots often paint images on canvas by executing sequences of brushstrokes and adjusting based on sensory feedback \cite{scalera2024history,lindemeier2013image,schaldenbrand2023frida}. Unlike digital image generation, a painting robot must master complex low-level control of deformable tools and fluids, force-sensitive manipulation of a brush against a surface, visual perception of the evolving artwork, and high-level planning of stroke sequences. These requirements make robot painting reproduction a significant challenge: the robot must achieve the same nuanced, variable brushstrokes as a human artist based on real-world physics. 

We ask: \emph{Given only a sequence of static images of an oil painting by an expert artist, can a robot infer corresponding control actions, such as trajectory, orientation, and applied force, to approximately reproduce the painting?} This setting is similar to art training exercises---can our robot reach the level of beginning art students who are able to replicate an oil painting? This is a challenging setting for robotics, as it simultaneously 1) requires precise low-level control to replicate strokes with soft brushes and wet paints; 2) requires high-level planning to compose multiple strokes; 3) assumes no access to any form of action demonstration data, either through teleoperation or tracking the brushes. 

One approach to tackling this challenge is \emph{learning a 2D dynamics model} for the oil painting domain. Given the current visual state of the canvas, and a parameterized action, the robot should predict the consequence of its brush action, i.e., the resulting state of the canvas. We conjecture that such a model can be trained through self-play. Once trained, the robot can integrate this model with model predictive control (MPC) applied within local stroke windows to infer actions to replicate each brushstroke,  plan multiple strokes to approximate a complex painting. Although learning neural dynamics models and combining them with MPC have become increasingly popular in different robotic tasks \cite{finn2017deep,shi2023robocook,wang2023dynamic,wu2023daydreamer,huang2022medor}, we believe this is the first attempt to leverage them for robot painting.  

We present \textbf{\algo} (\textbf{I}ntegrating \textbf{M}odel-Based \textbf{P}lanning with Le\textbf{A}rned Dynamic\textbf{S} Models for Robo\textbf{T}ic \textbf{O}il Painting Reproduction), a robotic system that combines learned dynamics models with force-sensitive control and model-based planning to approximately reproduce human oil paintings. Our contributions include:
\begin{itemize}
    \item Learned pixel dynamics model: We develop a neural network dynamics model of brushstrokes, trained on a dataset of robot self-play data. This model learns the complex relationship between the robot’s action parameters (end-effector trajectory, applied force) and the resulting stroke outcomes to predict stroke shape and appearance.
    \item Model predictive trajectory and force planning: We integrate the learned model into a model-based planning framework for single or consecutive stroke planning. Given a desired appearance of stroke(s), the MPC optimizer computes precise trajectories and forces to reproduce each brushstroke. 
    \item An integrated robot painting system: We implement a complete system on a 7-DoF robotic arm with a force-torque sensor. The system autonomously handles dipping the brush in paint, cleaning the brush between colors, and painting the strokes on a canvas.
\end{itemize}

We evaluate \algo on both individual strokes and full paintings drawn by expert human artists. Our results suggest that the learned model plus model-based planning approach yields strokes that closely match the appearances of the human brushstrokes. We demonstrate that the learned dynamics model can better replicate individual and overlaid strokes than the baseline methods. We also show closed-loop painting experiments in which the robot plans and paints a multi-stroke artwork.




%% file: 2-related.tex
\section{Related Work}
\subsection{Modern Robot Painting} The idea of automating art dates back centuries \cite{scalera2024history}. Harold Cohen's AARON in 1986 is widely considered the first modern robot painter with an open-loop programmed approach \cite{cohen1995further}. Recently, painting robots have employed closed-loop control. e-David is an industrial robotic arm equipped with a camera that paints with real brushes and continuously adjusts its strategy based on visual feedback \cite{lindemeier2013image,deussen2012feedback}. Another state-of-the-art system is the FRIDA series \cite{schaldenbrand2023frida,schaldenbrand2024cofrida,chen2025spline}, a collaborative painting robot that introduces a differentiable simulation for brushstrokes and a real2sim2real planning loop. FRIDA simulates how each stroke will appear and, during execution, periodically captures an image of the canvas to re-plan the remaining strokes, though FRIDA’s primary goal was not exact reproduction of existing artworks. These modern robotic painting systems emphasize sophisticated stroke planning algorithms, from high-level image segmentation and stroke sequencing \cite{lindemeier2013image} to optimization of stroke parameters in simulation \cite{schaldenbrand2023frida}, all aimed at achieving human-like painting results under robotic control. 

Recent advances in robotic painting leverage machine learning methods. Researchers have applied reinforcement learning (RL) to optimize paint coverage \cite{berrocal5396199reinforcement,tiboni2023paintnet} and stroke planning \cite{schaldenbrand2021content,lee2022scratch,jia2023sim}, used genetic algorithms for stroke planning \cite{aguilar2008robotic}, used imitation learning (IL) to leverage human demonstrations \cite{park2022robot,guo2022shadowpainter}, and employed deep generative models to synthesize and stylize brushstrokes \cite{wu2018brush,liu2021paint,gulzow2020recent,guo2024b}. Together, these approaches push robotic art beyond hand-designed programming toward more autonomous and creative behaviors. It is worth noticing that our problem setting is among the most challenging ones: We aim to closely approximate human experts' brushstrokes with soft brushes and wet pigments. Unlike in the RL settings, we do not rely on a painting simulator; unlike in the IL setting, we do not have access to human demonstrations.

\subsection{Planning with Learned Neural Dynamics Models}
Deep neural networks trained on interaction data have become a standard way to learn dynamics models for manipulation, with broad empirical success \cite{shi2023robocook,wang2023dynamic}. Such models can be trained directly in pixel space \cite{finn2016unsupervised,ebert2017self,ebert2018visual,yen2020experience,suh2020surprising} or in compact latent spaces that abstract observations \cite{watter2015embed,agrawal2016learning,hafner2019dream,hafner2019planet,schrittwieser2020mastering,wu2023daydreamer}. Beyond these representations, structured scene encodings improve modeling fidelity and generalization, including keypoints \cite{kulkarni2019unsupervised,manuelli2020keypoints,li2020causal,shenbab}, particles \cite{li2018learning,shi2022robocraft,zhang2024adaptigraph}, and mesh-based parameterizations \cite{huang2022medor}. For robot painting, a pixel-based dynamics model is a natural choice. Planning over learned dynamics is difficult due to their nonlinearity and nonconvexity. A prevalent strategy is online sampling-based optimization, most commonly cross-entropy methods (CEM) \cite{rubinstein2013cross} and Model Predictive Path Integral (MPPI) \cite{williams2017model}, as used by prior work for planning in manipulation tasks \cite{yen2020experience,ebert2017self,nagabandi2020deep,finn2017deep,manuelli2020keypoints,sacks2023deep,han2024model}. These methods require a large number of samples as the action dimension grows. Meanwhile, gradient-based trajectory optimization \cite{li2018learning,li2019propagation} is prone to local minima and non-smooth objectives.

%% file: 3-method.tex
\section{Method}
We describe the parameterization of brushstrokes, color prediction, the hardware system setup for oil painting, the architecture design and training of the dynamics model, the multi-step planning algorithm, and baseline methods for comparison.

\begin{figure}[b]
    \centering
    \includegraphics[width=0.75\columnwidth]{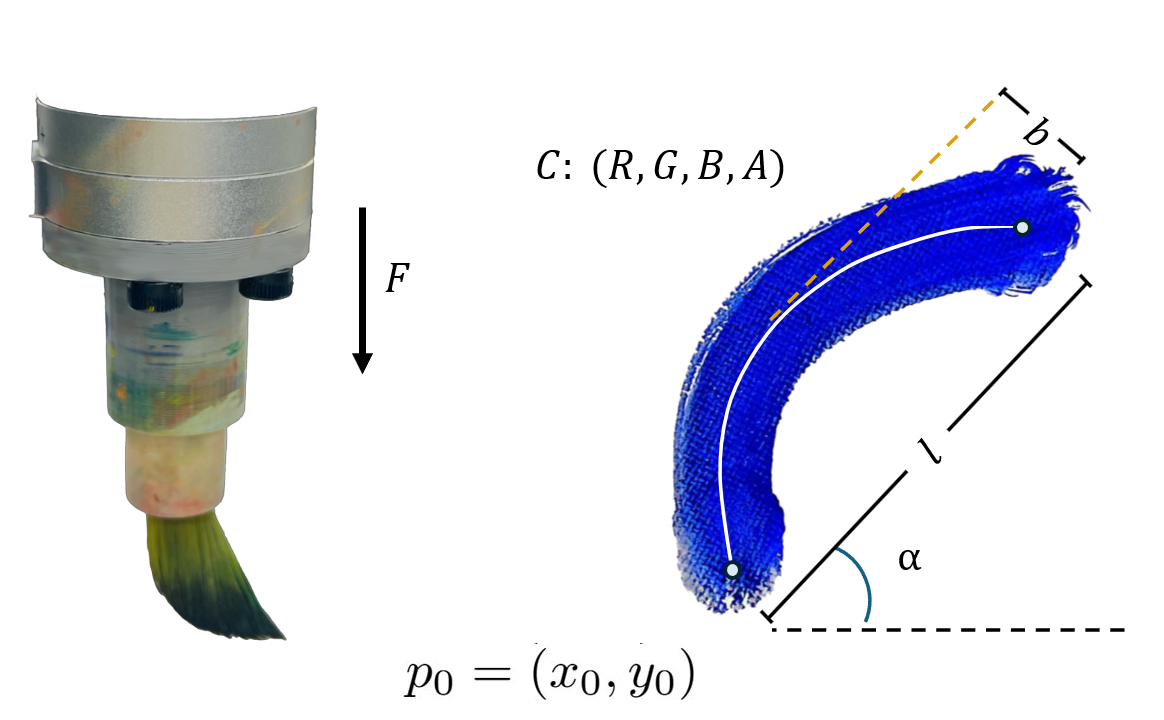}
    \caption{Action parameterization. Each brushstroke is represented by a quadratic Bézier curve, with a starting location $p_0=(x_0,y_0)$, length $l$, bend $b$, orientation $\alpha$, force $F$ (controls thickness), and color $C\in[0,1]^4$ (RGBA).}
    \label{fig:brush}
    \vspace{-0.2cm}
\end{figure}


\subsection{Brushstroke Model}
Although there are several possible ways to model brushstrokes \cite{guo2024b,wang2020robot,schaldenbrand2023frida,li2020differentiable}, oil paintings require curved strokes with precise force control. For a compact descriptive representation, we parameterize a brushstroke by six parameters, as shown in Fig.~\ref{fig:brush}: the starting location $p_0=(x_0,y_0)$ in image pixels, the stroke length $l$, a scalar bend $b$ that curves the stroke up or down, the in-plane orientation $\alpha$ (degrees), normal force $F$ that modulates deposited paint thickness and width, and the paint color $C=(R,G,B,A)\in[0,1]^4$. Compared to FRIDA \cite{schaldenbrand2023frida}, we replace their parameter $h$, which specifies how far the brush is pressed to the canvas, with the force parameter $F$, allowing for force-sensitive control. We use a separate module to predict $C=(R,G,B,A)$, and denote each brushstroke as Bezier [Fig.~\ref{fig:brush}] geometric control parameters:
\begin{equation}
    u = (p_0,\, l,\, b,\, \alpha,\, F),
\end{equation}
where we clip $u$ to task-specific bounds $\mathcal{U}$ for learning. This stroke representation cannot capture highly complex stroke patterns, such as serif-like endings or abrupt direction changes. Brushstroke execution is inherently stochastic and we account for this by explicitly controlling 
brush and paint conditions during evaluation.

\subsection{Color Prediction}
We separate color estimation from shape prediction, so the dynamics model focuses on stroke geometry. We run a nearest-neighbor search over a patch database of pigments: inside the stroke-difference mask, we form a representative feature and compare it to patch prototypes using a linear-RGB with Mahalanobis distance \cite{de2000mahalanobis}. Transparency is decided by a ratio test between base and target intensities. Additional details can be found in Appendix B.

\subsection{The Hardware System and Robot Control}
\label{sec:hardware}
\begin{figure}[t]
    \centering
    \includegraphics[width=0.9\columnwidth]{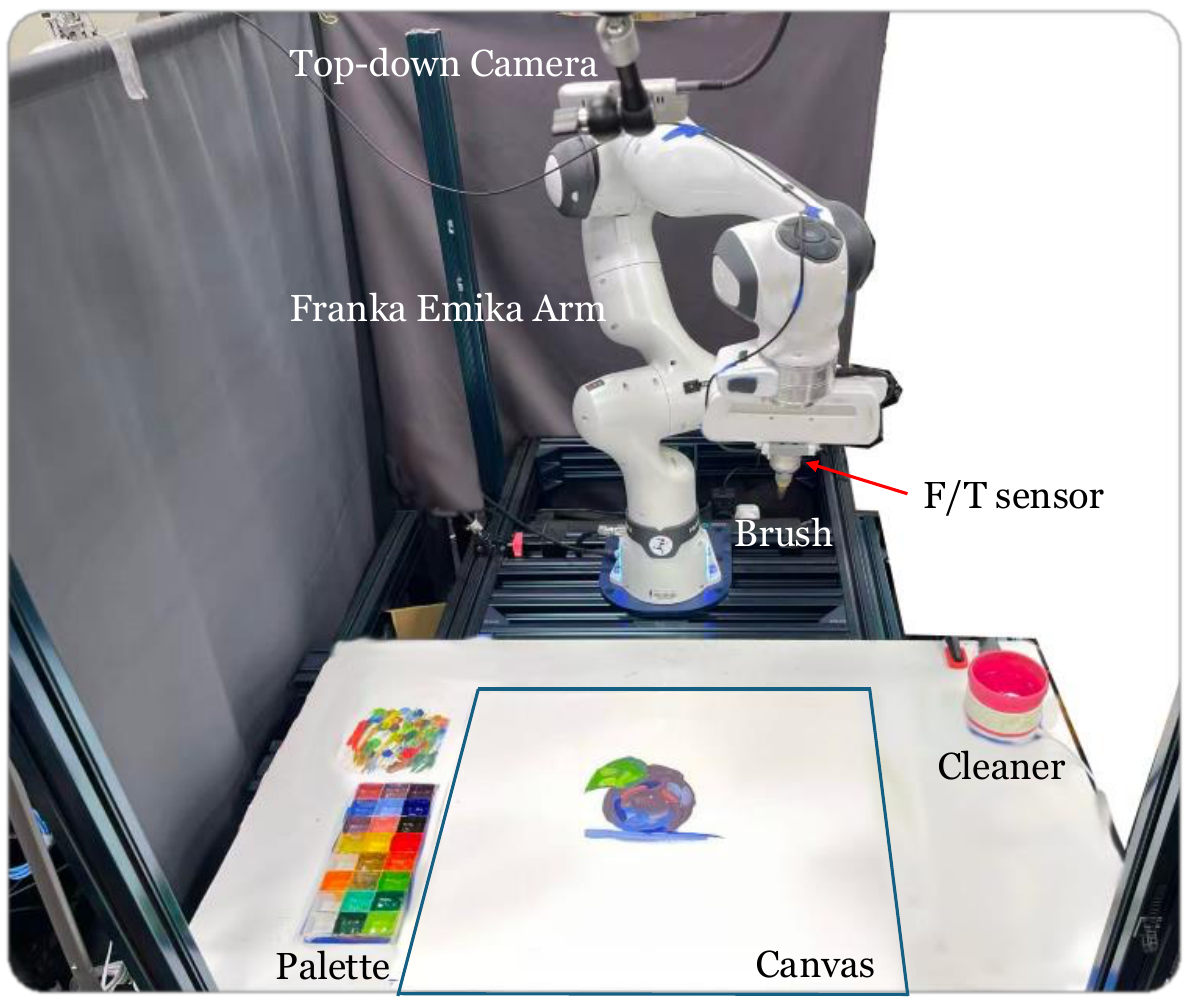}
    \caption{Hardware system setup of \algo. }
    \label{fig:setup}
    \vspace{-0.5cm}
\end{figure}

To execute brushstrokes on a canvas, we use a 7-DoF Franka Emika Panda equipped with a 6-axis force/torque sensor (Kunwei Tech Inc.) mounted at the tip using a 3D-printed mounter, as shown in Fig.~\ref{fig:setup}. Using an external force-torque sensor reduces dependence on robot-specific force estimation and improves reproducibility across manipulators. The force/torque sensor measures forces from $0.1N$ to $4N$, making it suitable for oil painting. A round synthetic-filament brush (Zen™ Series 43, size 8) trimmed to 5 cm is rigidly attached to the end effector and held perpendicular to the canvas. We control the arm via the Franka Control Interface in impedance mode and implement a force loop along the surface normal. At the beginning of each stroke, an outer normal-force admittance accumulates a smoothed feedforward force as:
\begin{equation}
F^{\mathrm{ff}}_{z,k+1}=F^{\mathrm{ff}}_{z,k}+k_f\,\mathrm{EMA}_\lambda[F^\star - F_{z,k}]\,\Delta t,
\end{equation}
where $k_f$ is the feedforward admittance gain, $F^\star$ is the desired normal force set by the stroke action’s $F$, $F_{z,k}$ is the measured normal force at step $k$,
$\mathrm{EMA}_\lambda[\cdot]$ is an exponential moving average with coefficient $\lambda$ defined as:
\begin{equation}
\bar{e}_{z,k} = \lambda\,\bar{e}_{z,k-1} + (1-\lambda)\,(F^\star - F_{z,k}),
\end{equation}
where $\bar{e}_{z,k}$ denotes the filtered force tracking error.
The controller injects force $\mathbf{F}_b^{\mathrm{ff}}=[0,0,F^{\mathrm{ff}}_z]^\top$ in the base frame via the Jacobian transpose:

\begin{equation}
{\tau}
={K}_p({q}_d-{q})-{K}_d\,\dot{{q}}
+{J}_p^{\!\top}\,s\,{F}^{{ff}}_{ee}.
\end{equation}
\noindent where $s$ is a scaling factor. Once the target contact force is established at the beginning of a stroke, the controller switches to impedance mode for the remainder of the stroke, maintaining the same vertical position. We found this sufficient in practice, as the canvas is approximately 
flat over the scale of a single stroke.
The canvas is an $18\times21$ inch board. Premixed pigments (24 colors) are provided in palette trays; a motorized water spinner cleans the brush between strokes. A top-down Intel RealSense D435 provides RGB observations; images are undistorted, homography-warped to the canvas plane, and center-cropped for learning. Additional details about the hardware setup are in Appendix A.

\subsection{The Pixel Dynamics Model and Training}
\label{sec:model}
\begin{figure*}[th!]
    \centering
    \includegraphics[width=0.85\textwidth]{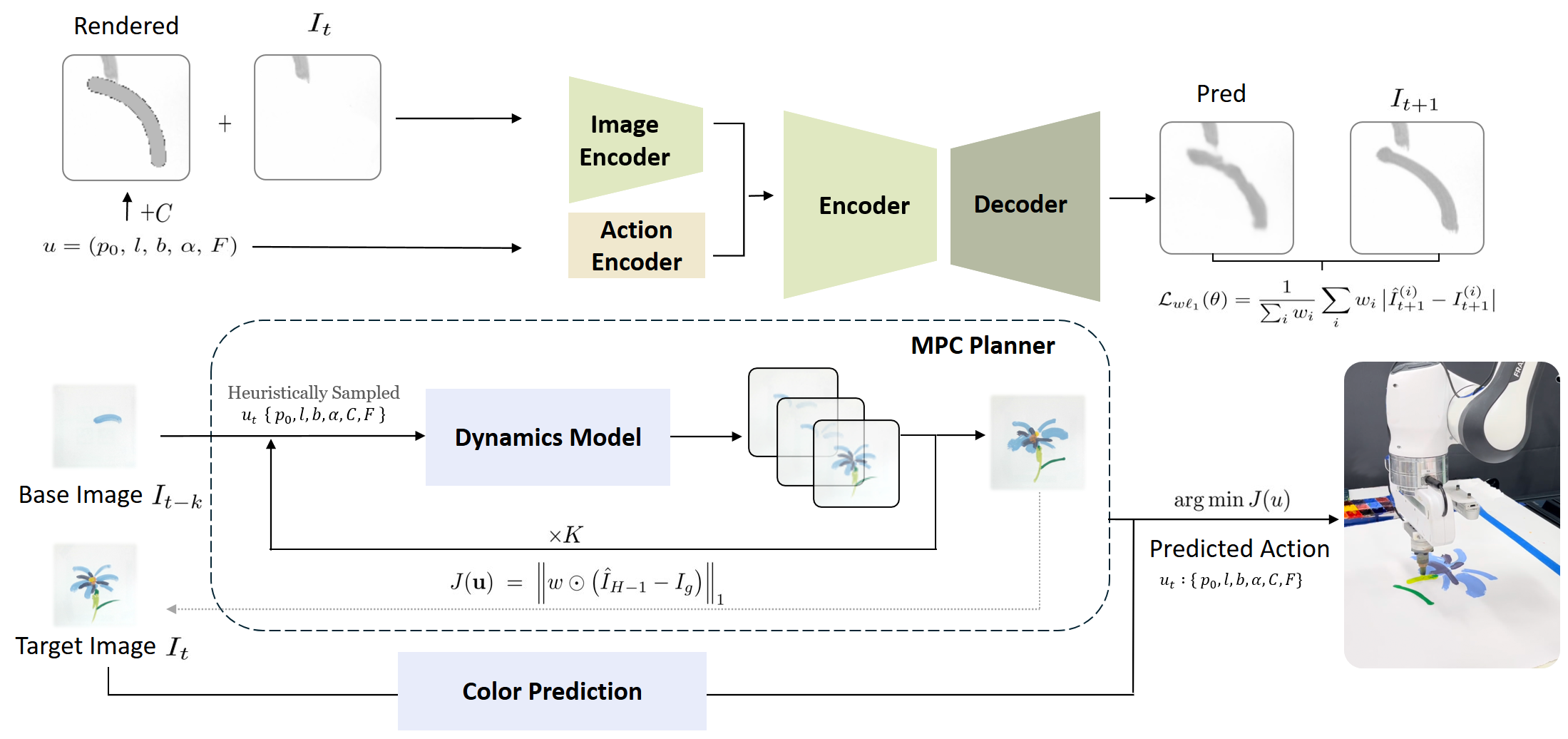}
    \caption{Overview of the learning and planning framework. Top: \algo-UNet's neural pixel dynamics model, which combines an image encoder and an action encoder to predict the effect of a stroke. The model is trained using a weighted $\ell_1$ loss. Bottom: To find one or more consecutive strokes between a base image and a target image, an MPC-based planner optimizes stroke parameters with a weighted $\ell_1$ image objective in a receding-horizon, closed loop.}
    \label{fig:method}
    \vspace{-0.4cm}
\end{figure*}

\paragraph{Pixel dynamics model}
The core challenge of this work is to design and train a forward dynamics model for oil painting. Given the current image observation $I_t$ and a candidate stroke action $u_t$, a differentiable dynamics model predicts the next observation:
\begin{equation}
\hat I_{t+1} = f_\theta(I_t, u_t).
\end{equation}

Architecture-wise, we employ an image encoder $\phi$ and an action encoder $\psi$ whose embeddings are fused by a decoder $\varphi$ to output $\hat I_{t+1}$. We pre-render stroke actions $u_t$ as an image $R_t$ using the color prediction $C_t$, and concatenate $R_t$ with $I_t$. Concretely,
\begin{equation}
\begin{aligned}
z_I &= \phi(\mathrm{Concat}(I_t, R_t)), \quad z_u = \psi(u_t),\\
\hat I_{t+1} &= \varphi\!\big(\mathrm{Concat}(z_I,\, z_u)\big).
\end{aligned}
\end{equation}

The neural network architecture is shown in Fig.~\ref{fig:method}. The encoder-decoder follows U-Net \cite{ronneberger2015u}. Each action $u_t$ is represented as a 6D vector of $(p_0(x), p_0(y), l, b, F, \alpha)$. 

The pre-rendered image $R_t$ bridges the modality gap between low-dimensional vector controls and high-dimensional raster states, enabling the network to perceive footprint, orientation, and thickness directly in image space. To map $(l,b,\alpha)$ into a continuous stroke centerline, we construct a quadratic Bézier curve with control points
\begin{equation}
\mathbf{q}_0 = p_0,\quad
\mathbf{q}_1 = p_0 + \tfrac{l}{2}\,\mathbf{t}_\alpha + b\,\mathbf{n}_\alpha,\quad
\mathbf{q}_2 = p_0 + l\,\mathbf{t}_\alpha,
\end{equation}
where $\mathbf{t}_\alpha=[\cos\alpha,\,\sin\alpha]^\top$ and $\mathbf{n}_\alpha=[-\sin\alpha,\,\cos\alpha]^\top$. The rendered stroke is then produced by rasterizing the curve with thickness $w(F)$ and compositing it onto the canvas, colored using $C_t$. The brush width $w(F)$ is obtained from a calibration curve that monotonically maps normal force to footprint width.

All the input images of the neural network are cropped around the target stroke and resized to $100\times100\times1$ grayscale patches.


\paragraph{Dataset and training}
The robot self-supervises to collect a dataset $\mathcal{D}=\{(I_t, u_t, I_{t+1})\}$ by uniformly randomly exploring over $(p_0,l,b,\alpha,F)$ while cycling a fixed color set; color-prediction runs use ground-truth color labels. The main objective is a weighted $\ell_1$ loss that emphasizes changed pixels near the stroke:
\begin{equation}
\mathcal{L}_{w\ell_1}(\theta)=\frac{1}{\sum_i w_i}\sum_{i} w_i \,
\big\lvert \hat I_{t+1}^{(i)} - I_{t+1}^{(i)} \big\rvert,
\end{equation}
with $w$ computed from a dilated difference mask between $I_t$ and $I_{t+1}$. We apply standard data augmentation (slight random cropping, rotation, and flipping) to augment the dataset. We train $f_\theta$ end-to-end with the Adam optimizer. Additional details of the dynamics model design and training can be found in Appendix B.

\subsection{Multi-Step Stroke Planning}
\label{sec:planning}
The learned forward dynamics model allows multi-step planning to complete a full painting. At test time, we plan in the action space with receding-horizon model predictive control (MPC). Starting from the current canvas observation $I_0$ and the goal image $I_g$, we optimize a sequence of $H=5$ actions and execute the first two before replanning. 
The horizon and execution step are chosen empirically to balance planning quality 
and computational cost. We then re-observe the canvas and re-optimize over the remaining planning horizon. Concretely, at each replanning cycle, we optimize a length-$H$ action sequence $\mathbf{u}=(u_0,\dots,u_{H-1})$ using a weighted image-space objective:
\begin{equation}
\begin{aligned}
J(\mathbf{u}) \;=\; \left\lVert w \odot \bigl(\hat I_{H-1}- I_g \bigr) \right\rVert_1,
\quad \\
\text{s.t.}\;\; \hat I_{t+1}=f_\theta(\hat I_t,u_t),\; \hat I_0=I_0
\end{aligned}
\end{equation}

where $w$ up-weights goal-relevant pixels (stroke masks). We draw $H$ heuristic initializations through a PCA–RANSAC multi-way splitter, perform adaptive sampling and elite-set updates via MPPI \cite{williams2017model}, clip to $\mathcal{U}$, select the optimized candidate, execute $u_t^\star$, and re-plan (Alg.~1). This closed-loop, sample-and-refine procedure is robust to model error and naturally incorporates action bounds and force limits. More details can be found in Appendix C.

\begin{algorithm}
\caption{Receding-Horizon MPC with Adaptive Covariance}
\begin{algorithmic}[1]
\Require Current observation\ $I_0$, goal image $I_g$, dynamics model $f_\theta$, cost $c$ ( $L_{w\ell_1}$), candidates $K$, elite ratio $\rho$, temperature $\beta$, horizon $H$, EMA $\texttt{ema}$
\Ensure Action sequence $\mathbf{U}^\star = (u_0^\star,\dots,u_{H-1}^\star)$
\State $\mathbf{U} \gets \textproc{InitFromMasks}(I_0,I_g)$ \Comment{splitter $\rightarrow$ initial $H$}
\For{$t = 0, \dots, H{-}1$} \Comment{MPC recedes, feeds next step}
 \For{$\text{iter}=1,\dots,M$} \Comment{sample-and-refine}
    \State $\mathcal{A} \gets \{\mathbf{U}\}$ \Comment{null particle (no noise)}
    \State $\mathcal{A} \gets \mathcal{A} \cup \textproc{SampleAdaptive}(\mathbf{U}, K{-}|\mathcal{A}|)$
    \ForAll{$\tilde{\mathbf{U}} \in \mathcal{A}$}
      \State $\hat I_{H} \gets \textproc{Rollout}(f_\theta, I_t, \tilde{\mathbf{U}})$
      \State $J(\tilde{\mathbf{U}}) \gets \lVert w \odot (\hat I_{H} - I_g) \rVert_1$ \Comment{terminal $L_{w\ell_1}$}
    \EndFor
    \State $\mathcal{E} \gets \text{top-}K_e \text{ by } J$ \Comment{$K_e=\lfloor\rho K\rfloor$}
    \State $\mathbf{U} \gets \texttt{ema}\cdot\sum_{\tilde{\mathbf{U}}\in\mathcal{E}}\textproc{Softmax}_{\beta}(-J)\,\tilde{\mathbf{U}} \;+\; (1{-}\texttt{ema})\cdot \mathbf{U}$
    \State $(\Sigma_t,\sigma_t)_{t} \gets \textproc{CMAUpdate}(\mathcal{E})$ \Comment{update per-timestep cov/step-size}
    \State $\mathbf{U} \gets \textproc{Clip} \ \mathbf{U}\text{ to action bounds}$
  \EndFor
  \State $I_{t+1} \gets \textproc{execute}(I_t, u_t^\star)$ \Comment{Apply to system / obtain next obs.}
\EndFor
\State \Return $\mathbf{U}^\star$
\end{algorithmic}
\end{algorithm}

\subsection{Baseline Methods}
\label{sec:baselines}
We denote our main model as \algo-UNet. Now we introduce several ablations, which are variants of our method or from related work, including:

\noindent \textbf{\algo-LR:} Instead of using a U-Net backbone, this ablation fits a linear regressor that maps the force $F$ into the stroke thickness, and renders $u = (p_0, l, b, \alpha, F)$ directly to an alpha map, which is then composited with the base image and a constant stroke color. This is similar to the method used in Spline-FRIDA \cite{chen2025spline}, where \(F\!\in\![0,1]\) are mapped to a global stroke thickness $\tau$:
\begin{equation}
\tau(F)=\operatorname{softplus}(a F+\beta)+\varepsilon
\end{equation}
and a parabolic trajectory represented by  $(p_0, l, b, \alpha)$ is rasterized by a sequence of spheres with $\tau$ as radius.

\noindent \textbf{Heuristics-only:} For single-step stroke planning, we skeletonize the frame-to-frame difference mask and use the two farthest skeleton endpoints (for $p_0$ and $l$), their direction (for $\alpha$), and the skeleton’s maximum signed normal offset (for $b$), with a fixed force $F = 0.5$.

\noindent \textbf{FRIDA-CNN:} FRIDA \cite{schaldenbrand2023frida} was not developed to exactly reproduce brushstrokes. FRIDA-CNN only uses FRIDA's \texttt{param2stroke} model, a convolutional neural network predicting a stroke occupancy field from $(l, b, F)$, and is a variant of our method. The stroke occupancy is transformed and blended using $(p_0, \alpha,C)$ over the base image.

Additional details of the baselines are in Appendix B. Note that, unlike \algo, all baseline methods operate \emph{without access to the underlying state context} (i.e., the base image). They \emph{predict only the incremental stroke rather than the next full state image}. To approximate state prediction, we overlay the predicted increment onto the ground-truth base image. In evaluation, this places \algo-UNet at a disadvantage in terms of the weighted $\ell_1$ loss, as it must predict the entire next image. 

%% file: 4-results.tex
\section{Experiments and Results}
Our experiments are designed to address three key questions: $\mathbf{\mathcal{Q}1}$. How much data is needed to train a dynamics model in our case? $\mathbf{\mathcal{Q}2}$. Can \algo reproduce expert human artists' strokes with low error and high visual similarity? $\mathbf{\mathcal{Q}3}$. Can \algo enable closed-loop, multi-step planning to approximate a given oil painting?

\subsection*{$\mathbf{\mathcal{Q}1}$. Can \algo's pixel dynamics model be trained to predict the stroke effects with high fidelity?} 
\begin{figure}
    \centering
    \includegraphics[width=0.9\linewidth]{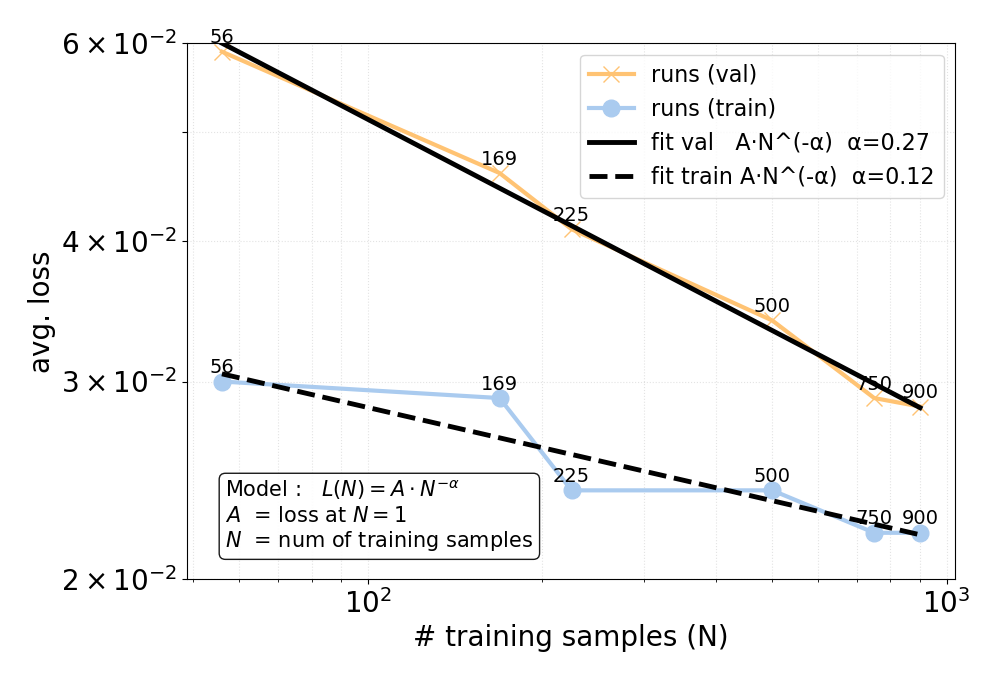}
    \caption{Training and evaluation $\mathcal{L}_{\ell_1}$ (unweighted) of \algo's dynamics model vs. number of training samples (log–log scale). }
    \label{fig:sample_efficiency}
    \vspace{-0.3cm}
\end{figure}

As shown in Fig.~\ref{fig:sample_efficiency}, as the number of training samples increases, the dynamics model achieves lower prediction error when predicting the next canvas state. The validation set is fixed at 10\% of the total dataset size. We observe that with a reasonable amount of self-play data ($n=900$), the model can achieve a prediction with unweighted $\mathcal{L}_{\ell_1} = 0.0285$.

\subsection*{$\mathbf{\mathcal{Q}2}$. Can \algo reproduce human individual strokes with lower error and higher visual similarity than baseline methods?} 
We compare \algo-UNet with the baselines. Since real brush-paint interaction is inherently stochastic, we explicitly controlled the execution conditions in the final reported experiments. In particular, we matched the brush state and pigment condition for each stroke across baselines as close as possible, in order to reduce variability across trials and improve fairness in quantitative comparisons. 

First, we collected two datasets for evaluation. The first dataset (``Single Strokes'') comes from five trained human artists, who are experts in oil painting. Every artist contributed 12 separate brushstrokes that reflected their unique style. The second dataset (``Overlaid Strokes'') was produced by the robot, which freely painted 50 random strokes that were layered on top of each other. Note that these datasets are not used for training.

Then we integrate the learned dynamics model with the multi-step planning algorithm described in Sec.~\ref{sec:planning}. We first set the planning horizon to be 1 step. We measure the accuracy in two stages: the planning stage and the execution stage. At the planning stage, the 1-step planning algorithm infers the stroke actions, then we use \algo and baselines to predict and render the strokes, and compare the rendered strokes with ground-truth strokes. At the execution stage, we let the robot execute the stroke action, take a photo of the canvas, and compare that with the ground truth. 

We use two metrics for evaluation. The first one was the weighted $\ell_1$ loss used as the objective in learning and planning. The second one was the popular LPIPS (Learned Perceptual Image Patch Similarity) metric \cite{zhang2018unreasonable}, which is a perceptual metric that measures the similarity between two images based on deep neural network feature embeddings, aligning better with human visual judgments than pixel-wise differences. This metric is particularly useful for the execution stage due to the differences in canvas color and lighting conditions.

\begin{table}[]
    \centering
    \resizebox{\linewidth}{!}{
    \begin{tabular}{c|cccccc}
    \toprule
    Stage & Planning & \multicolumn{2}{c}{Execution}  & Planning & \multicolumn{2}{c}{Execution}\\ 
    Loss & $\mathcal{L}_{w\ell_1}\downarrow$ & $\mathcal{L}_{w\ell_1}\downarrow$ & LPIPS$\downarrow$ & $\mathcal{L}_{w\ell_1}\downarrow$ & $\mathcal{L}_{w\ell_1}\downarrow$ & LPIPS$\downarrow$ \\
    \midrule
    Test Dataset & \multicolumn{3}{c}{Artist \#1} & \multicolumn{3}{c}{Artist \#2}\\
    Heuristics-only & N/A & 0.0541 & 0.2001& N/A &0.0626 & 0.2001 \\
    FRIDA-CNN& 0.0215 & 0.0617&0.2736 & 0.0318&0.0508&0.2348 \\
    \algo-LR&  0.0229&0.0718 &0.2511 & 0.0242&0.0434&0.1711 \\
    \algo-UNet&  \textbf{0.0197}&  \textbf{0.0506}&\textbf{0.1874}& \textbf{0.0225}&\textbf{0.0413}&\textbf{0.1675} \\
    \midrule
    Test Dataset & \multicolumn{3}{c}{Artist \#3} & \multicolumn{3}{c}{Artist \#4}\\
    Heuristics-only & N/A &0.0834 & 0.2455& N/A &0.0556 &  0.2246\\
    FRIDA-CNN &0.0278 & 0.1062 & 0.3495& 0.0256 & 0.0362 & 0.2623 \\
    \algo-LR & \textbf{0.0202} & \textbf{0.0721}& 0.2622&0.0271&\textbf{0.0331 }& 0.2338 \\
    \algo-UNet & 0.0257&0.0745  &\textbf{0.2236} & \textbf{0.0234} &  0.0346 & \textbf{0.1621}  \\
    \midrule
    Test Dataset & \multicolumn{3}{c}{Artist \#5} & \multicolumn{3}{c}{Overlaid}\\
    Heuristics-only & N/A & 0.0574& 0.2017& N/A & 0.0940&0.2554 \\
    FRIDA-CNN &0.0264 &0.0515&  0.2781&0.0288 & 0.1034&0.2645\\
    \algo-LR & 0.0230&\textbf{0.0402}&  0.1869& 0.0263 & 0.0929&0.2473\\
    \algo-UNet &\textbf{ 0.0199}&0.0457& \textbf{0.1633}&\textbf{0.0258} & \textbf{0.0907}&\textbf{0.2209}\\
    \bottomrule
    \end{tabular}}
    \caption{Learned dynamics models' planning and execution performance in terms of $\mathcal{L}_{w\ell_1}$ and LPIPS losses. All methods were evaluated under controlled brush/paint state conditions.}
    \label{tab:ind_strokes}
    \vspace{-0.4cm}
\end{table}

\begin{figure}[tbh!]
    \centering
    \includegraphics[width=1\linewidth]{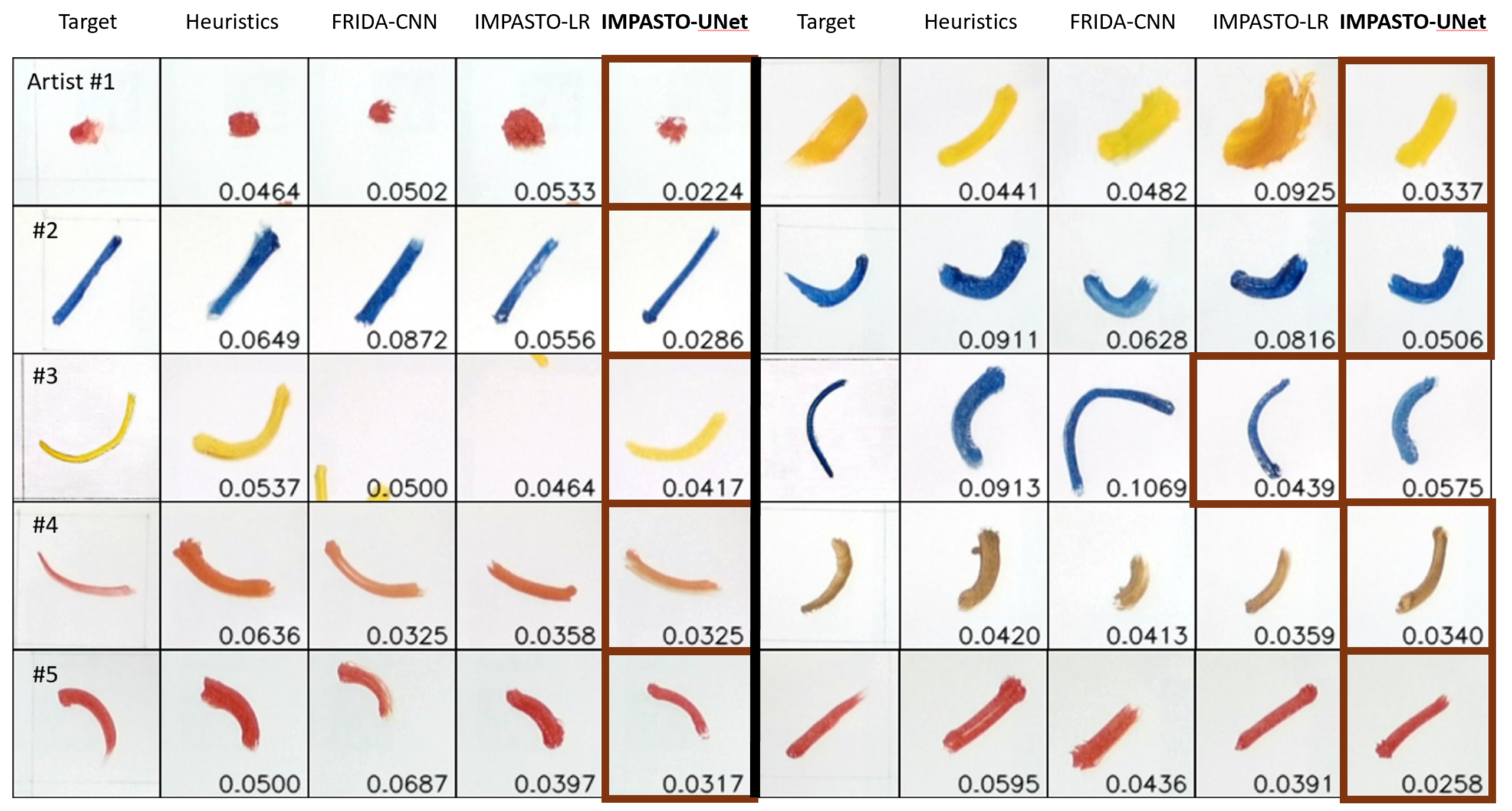}
    \caption{Target brushstrokes from five human artists (two examples per artist) and the strokes reproduced by the robot using different methods. The numbers shown are the weighted $\ell_1$ loss between the target and the painted strokes. Instances with the best performances are highlighted with bold borders. Overall, \algo-UNet approximated human brushstrokes with lower error and higher visual similarity.}
    \label{fig:ind_strokes}
    \vspace{-0.5cm}
\end{figure}

\begin{figure}
    \centering
    \includegraphics[width=0.9\linewidth]{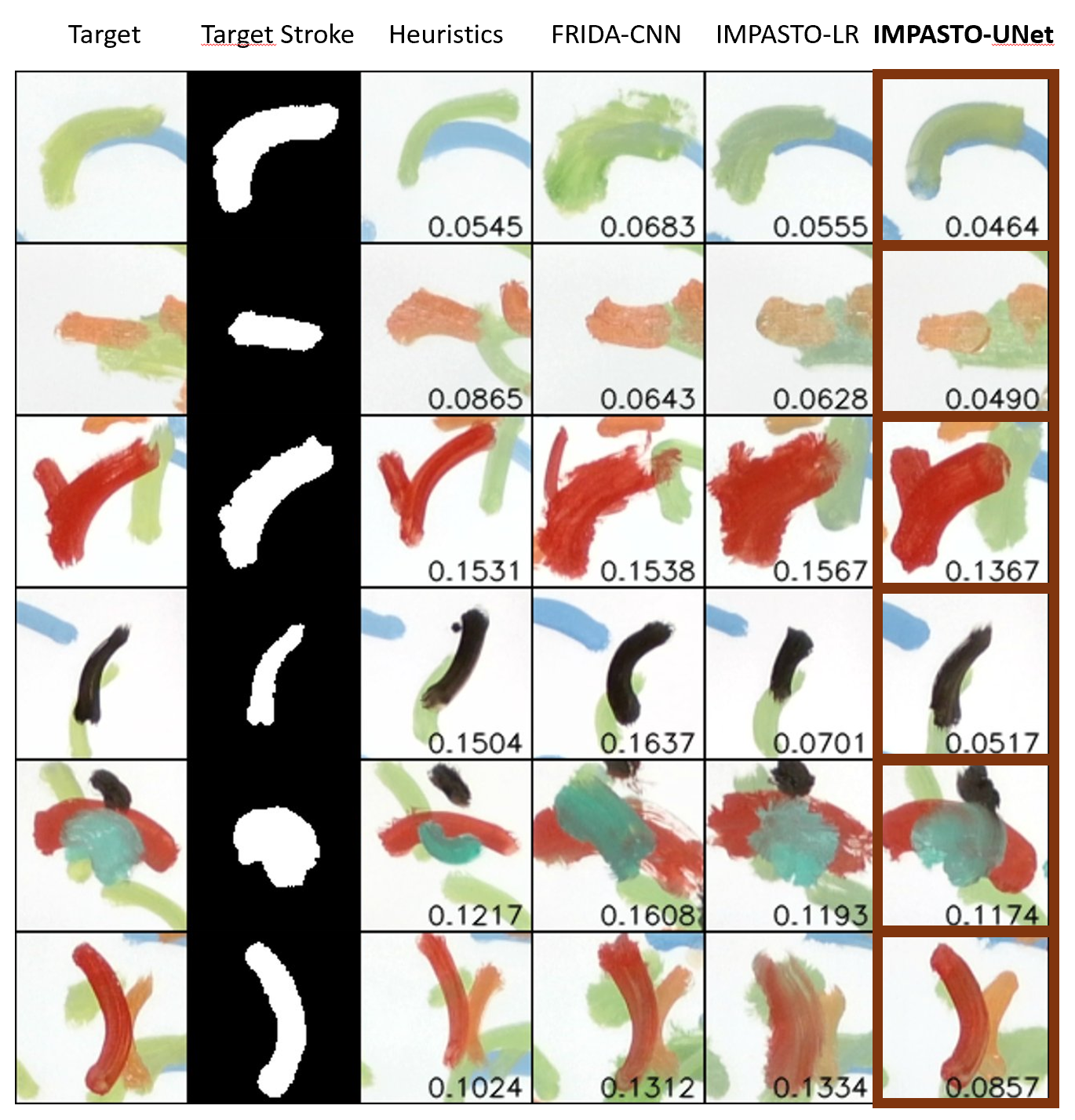}
    \caption{Qualitative results showing the target strokes from overlaid strokes and the strokes painted by the robot using different methods. Instances with the best performances are highlighted with bold borders. The difference with Fig.~\ref{fig:ind_strokes} is that the base images (canvas states) already have painted strokes. This requires the dynamics models to make predictions with low prediction errors given the noisy background. The numbers shown are the weighted $\ell_1$ loss. Note that the loss is \emph{only} calculated around the target stroke area. \algo-UNet is more accurate in reproducing brushstrokes given the noisy base images.}
    \label{fig:overlaid_strokes}
    \vspace{-0.2cm}
\end{figure}

The results are shown in Table~\ref{tab:ind_strokes}. The visualization of results can be found in Fig.~\ref{fig:ind_strokes} and Fig.~\ref{fig:overlaid_strokes}. Our U-Net-based model consistently performs better than baselines in terms of LPIPS metrics in all cases, although the linear regression variant performs slightly better in terms of the weighted $\ell_1$ loss for some artists. On average across all artists' data, \algo-UNet reduces planning $\mathcal{L}_{w\ell_1}$ loss by 16.45\% vs. FRIDA-CNN and 5.28\% vs. \algo-LR; execution $\mathcal{L}_{w\ell_1}$ loss by 19.48\%, 5.33\%, and 21.21\% vs. FRIDA-CNN, \algo-LR, and Heuristics-only; and LPIPS by 35.36\%, 18.21\%, and 15.68\%, respectively. Notably, although our dynamics model was trained using self-supervised play data, it performed reasonably well in reproducing brushstrokes from different human artists, showing generalization across human stroke styles. 

\algo also performs better than all the baselines in the ``Overlaid'' dataset, suggesting that it can make predictions with lower prediction errors given noisy base images. \algo-UNet reduces planning $\mathcal{L}_{w\ell_1}$ loss by 10.42\% vs. FRIDA-CNN and 1.90\% vs. \algo-LR; execution $\mathcal{L}_{w\ell_1}$ loss 12.28\%, 2.37\%, and 3.51\% vs. FRIDA-CNN, \algo-LR, and Heuristics-only; and LPIPS 16.48\%, 10.68\%, and 13.51\%.

\subsection*{$\mathbf{\mathcal{Q}3}$. Can \algo enable closed-loop, multi-step planning to approximate complex oil paintings?} 
\begin{figure*}[tbh!]
    \centering
    \includegraphics[width=1\linewidth]{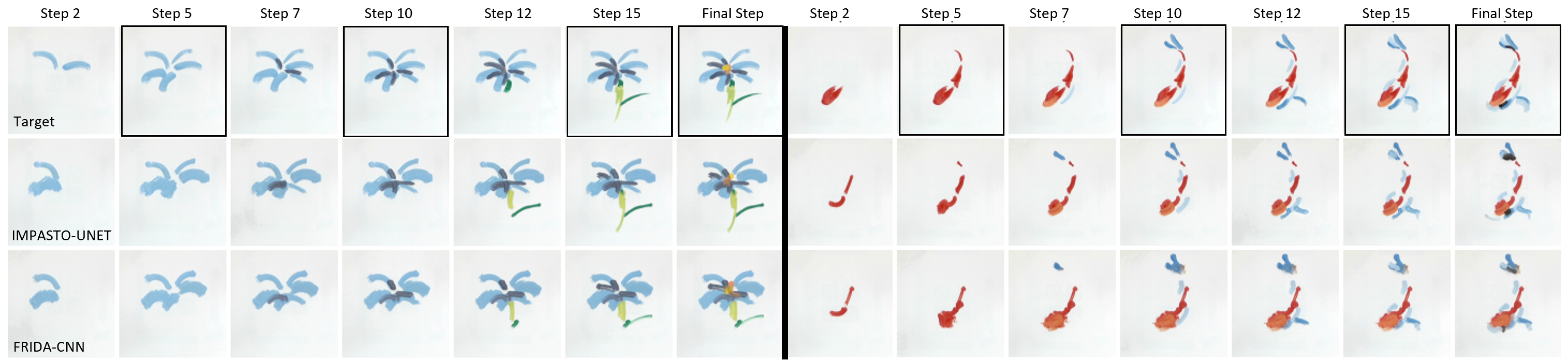}
    \caption{Qualitative results showing the target paintings and the painting produced by the robot using \algo vs. FRIDA. The planning horizon was set to $H = 5$, and the framed images are the targets for MPC within each prediction window. \algo can approximate oil paintings with better details in terms of forces and shapes, as shown by quantitative results in Table~\ref{tab:painting_strokes}.}
    \label{fig:painting_strokes}
    \vspace{-0.5cm}
\end{figure*}

Next, we show that \algo can integrate long-horizon, multi-step planning to approximate oil paintings with many strokes. We first ask an artist to paint two oil paintings (``Flower'' and ``Fish''), consisting of 17 strokes and 18 strokes, respectively. We then set the planning horizon to be $H = 5$, i.e., step 5, 10, 15, and the final images are used as target images for receding-horizon planning within each prediction window. We compare \algo with FRIDA-CNN, the best performing baseline other than \algo's own variant, using the same metrics. The quantitative results are shown in Table~\ref{tab:painting_strokes}, and the visualizations are in Fig.~\ref{fig:painting_strokes}. \algo performs better in multi-step planning, thanks to its better performance in reproducing individual brushstrokes. On average, \algo-UNet reduces planning $\mathcal{L}_{w\ell_1}$ loss by 24.01\% vs. FRIDA-CNN; execution $\mathcal{L}_{w\ell_1}$ loss by 10.55\% vs. FRIDA-CNN; and LPIPS by 14.63\%.  

\begin{figure}[tbh!]
    \centering
    \includegraphics[width=1\linewidth]{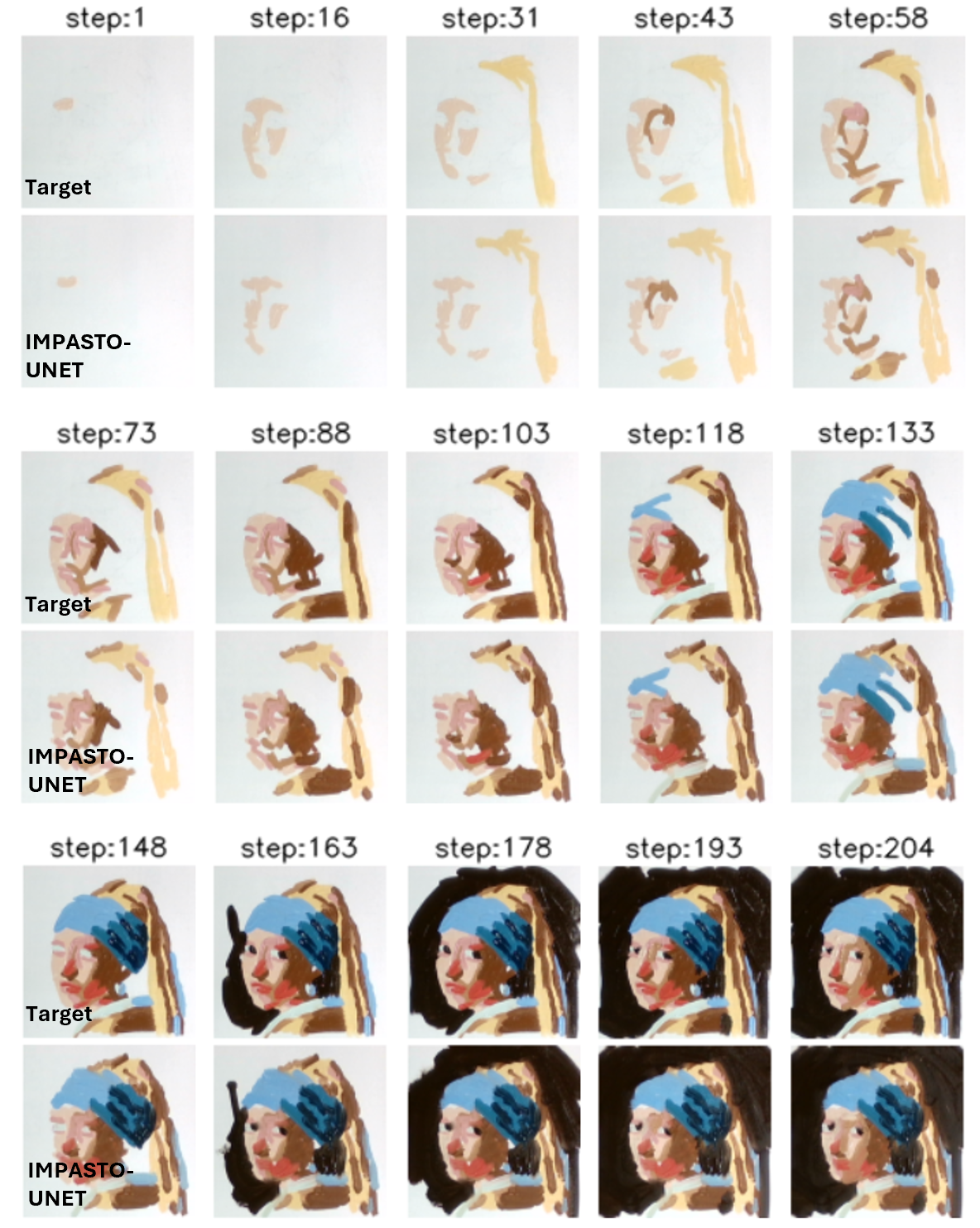}
    \caption{\algo was able to closely approximate a complex oil painting with 204 strokes (not all steps are shown). }
    \label{fig:final_demo}
    \vspace{-0.6cm}
\end{figure}

Finally, as shown in Fig.~\ref{fig:final_demo}, \algo was able to approximate a rather complex painting with high visual similarity, which requires 204 steps. More examples and the robot painting process can be found in Appendix D and the supplemental video.

\begin{table}[]
    \centering
    \resizebox{\linewidth}{!}{
    \begin{tabular}{c|cccccc}
    \toprule
    Test Data & \multicolumn{3}{c}{Flower} & \multicolumn{3}{c}{Fish}\\
    Stage & Planning & \multicolumn{2}{c}{Execution}  & Planning & \multicolumn{2}{c}{Execution}\\ 
    Loss & $\mathcal{L}_{w\ell_1}\downarrow$ & $\mathcal{L}_{w\ell_1}\downarrow$ & LPIPS$\downarrow$ & $\mathcal{L}_{w\ell_1}\downarrow$ & $\mathcal{L}_{w\ell_1}\downarrow$ & LPIPS$\downarrow$ \\
    \midrule
    FRIDA-CNN &0.0597 &0.0501&0.1619 & 0.0519 & 0.0523 & 0.1204 \\
    \algo-UNet &\textbf{0.0427} &\textbf{0.0464} &\textbf{0.1293} &\textbf{0.0421} &\textbf{0.0452} & \textbf{0.1117}\\
    \bottomrule
    \end{tabular}}
    \caption{Multi-step stroke planning and execution performance. Planning horizon $H = 5$. }
    \label{tab:painting_strokes}
    \vspace{-0.5cm}
\end{table}

%% file: 5-conclude.tex
\section{Conclusions}
An oil painting robot must address fine-grained control of deformable tools and fluid dynamics, execute force-sensitive brush-surface interactions, perceive the evolving canvas state, and solve high-level planning of stroke sequences. These requirements make robotic oil painting an inspiring and ambitious challenge for robotics research. We present \algo, a robot painting system that integrates learned dynamics models and model-based planning to approximate oil paintings. We demonstrate that \algo can approximate human artists' brushstrokes and artworks. Through these contributions, we advance robotic painting by fusing learning and control: the robot gains an understanding of brush dynamics from data and uses it in a principled planning framework. \algo represents a step toward robots that can paint with the finesse of a human artist, manipulating real brushes and paints. 

%% file: 9-appendix.tex
\FloatBarrier
\setcounter{figure}{0}
\renewcommand{\thefigure}{\Alph{section}\arabic{figure}}
\section*{Appendix A. Robotic Hardware System}
\begin{figure}
    \centering
    \includegraphics[width=1\linewidth]
    {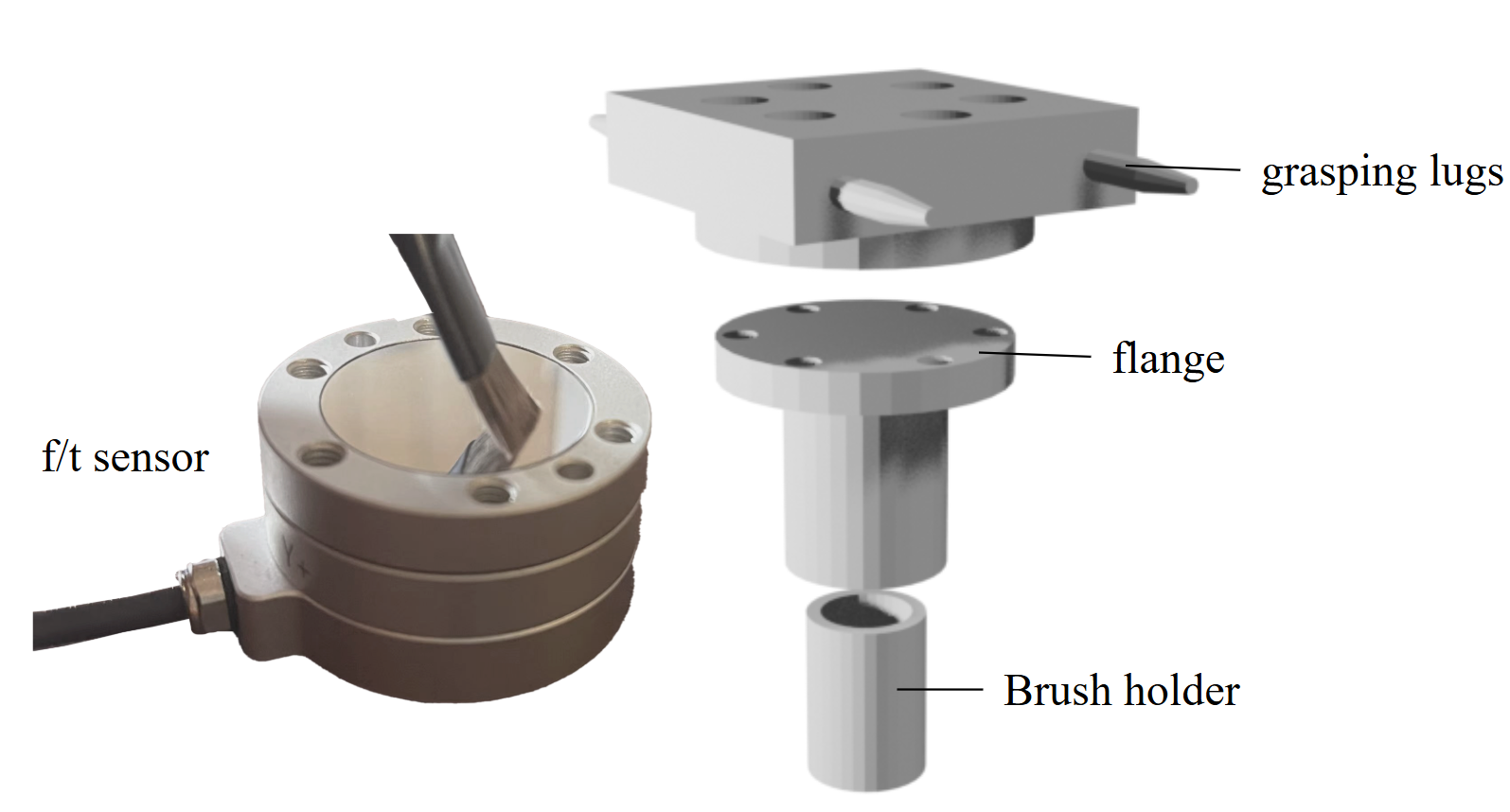}
    \caption{The f/t sensor and the mounter.}
    \label{fig:sensor}
\end{figure}

We use a 7-DoF Franka Emika Panda equipped with a 6-axis force/torque sensor (Kunwei Tech Inc.) mounted at the tip, as shown in Fig.~\ref{fig:sensor}. The mount is designed and 3D-printed, with a grasping lugs for the Franka's parallel grippers, a flange for the f/t sensor, and a brush holder for the oil painting brush. To reduce common-mode noise and mains interference, we ground the acquisition chain by connecting a grounding wire to the data-acquisition module’s adapter. Sensor readouts are first bias-corrected (offset removal) at the beginning of each primitive skill and then smoothed with an exponential moving average (EMA):
\begin{equation}
\tilde{x}_t = x_t - \hat{b},
\qquad
y_t = \alpha\,\tilde{x}_t + (1-\alpha)\,y_{t-1},
\label{eq:ema}
\end{equation}
\noindent where
$x_t$ denotes the raw force/torque reading at time step $t$;
$\hat{b}$ is the per-primitive bias (static offset) estimated before each primitive skill
from a short stationary window collected right before the primitive starts;
$\alpha \in (0,1]$ is the EMA coefficient.

The RGB observations of the canvas are captured at $1920\times1080$ by a single Intel RealSense D435 camera mounted above the canvas. 

Dynamics model training uses Distributed Data Parallel (DDP) on NVIDIA GPUs, including RTX A5000, RTX 3090, and A40. Deployment uses NVIDIA RTX 4090 GPUs. 

\section*{Appendix B. Learning the Dynamics Models}
\begin{figure*}
    \centering
    \includegraphics[width=0.9\linewidth]{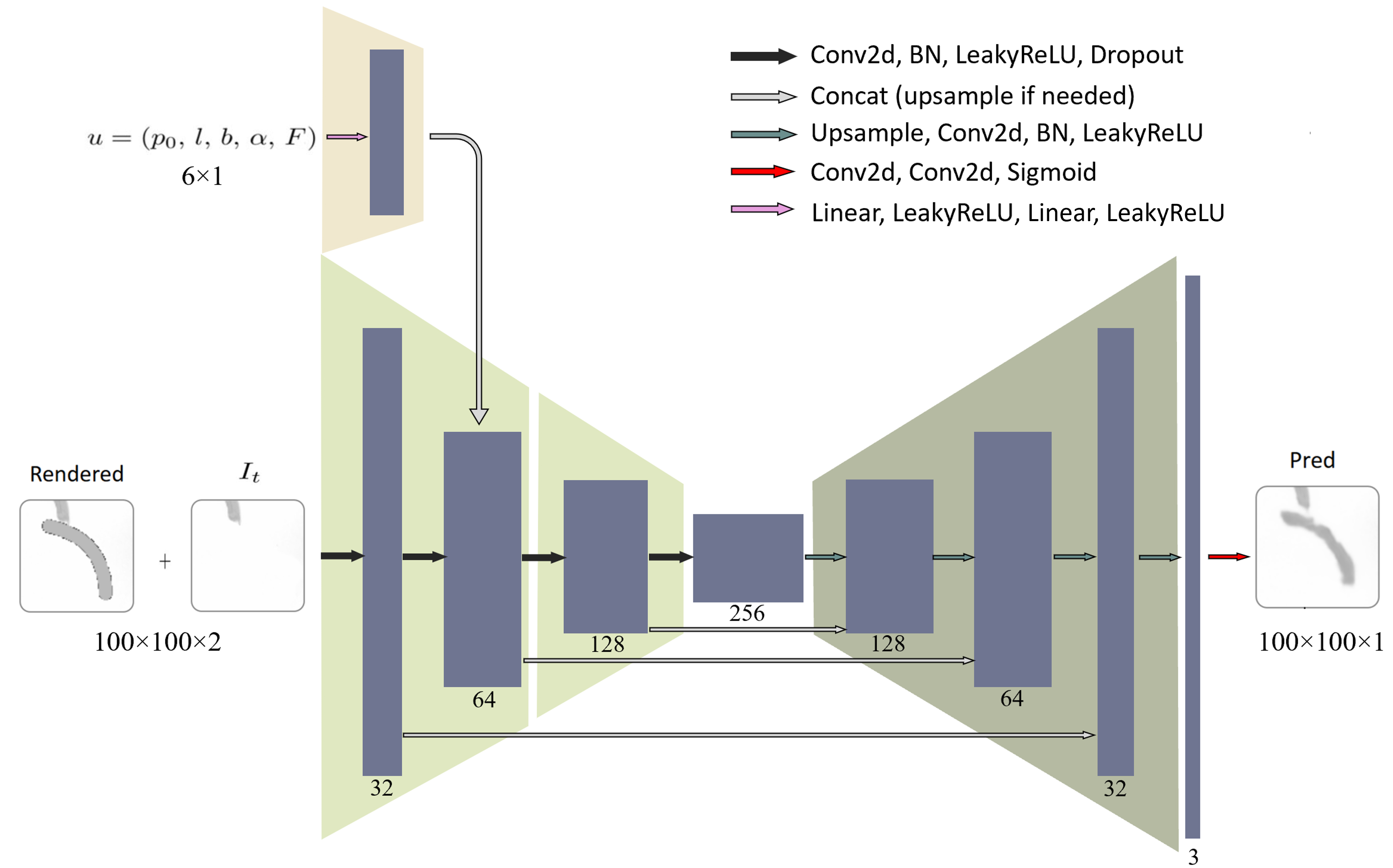}
    \caption{The neural network architecture for the pixel dynamics model.}
    \label{fig:nn}
\end{figure*}
The detailed neural network architecture is shown in Fig.~\ref{fig:nn}, where we use a U-Net–style encoder–decoder that fuses image evidence with action parameters. We train with the Adam optimizer (learning rate $5\times10^{-4}$, weight decay $10^{-3}$) using a batch size of 120 for 1000 epochs. The learning rate follows a multi-step schedule with milestones every 100 epochs, each decayed by a factor of $0.75$.
 We render each candidate stroke as a rasterized, disk-stamped footprint. Along the centerline—either a straight segment or a quadratic arc, depending on the absolute bend—we place $N$ uniformly spaced centers and stamp filled circles at each center. The circle radius follows a force–to–width power law:
 \begin{equation}r(F)=r_{\min}+k\,F^{\gamma},\end{equation} where $F$ is the applied force, $r_{\min}$ is a baseline radius, $k$ is a scale factor, and $\gamma>0$ controls the nonlinearity. Colors are treated as grayscale for simplicity: we emit three identical RGB channels (no hue variation) for compatibility with RGB loaders, and set the alpha channel to fully opaque. After rendering all circles, we resize the canvas to $100\times100$.

\subsection*{Baseline Implementation Details}

The Heuristic-only method recovers a stroke’s geometry directly from a binary mask without learning. We skeletonize the mask via iterative, topology-preserving thinning to obtain a 1\,-pixel-wide centerline, then identify endpoints as skeleton pixels with exactly one 8-neighborhood neighbor (degree$=1$). With endpoints \(p_0,p_1\), we define a local frame along the chord \(p_0\!\to\!p_1\), project skeleton points onto the tangent--normal basis, and estimate $b$ (bend) as the signed maximal lateral deviation (thresholded to suppress noise); $l$ (length) is \(\lVert p_1-p_0\rVert\) and $\alpha$ (angle) is the chord heading.

FRIDA-CNN maps a 3D geometric parameter vector $(l,b,F)$ to a grayscale stroke mask. A \texttt{BatchNorm} layer followed by a lightweight MLP produces a flattened $100{\times}100$ field, which is reshaped and refined by a small convolutional layer, yielding $M\in[0,1]^{100\times100}$. 
Since $M$ encodes geometry without translation and rotation, we transform $M$ using the remaining 3D geometric parameter $(p_0(x_0), p_0(y_0),\alpha)$.

Baselines FRIDA-CNN, \algo-LR, and Heuristics predict only the incremental stroke, rather than the full next state image as \algo-UNet does. To place the stroke on a base image $B\in[0,1]^{100\times100}$ with per-sample grayscale value $c\in[0,1]$ and opacity $a\in[0,1]$, we form a convex combination controlled by an effective opacity $A_{\mathrm{eff}}=\mathrm{clip}(M\cdot a,0,1)$:
\begin{equation}
I_{\mathrm{out}} \;=\; B\cdot (1 - A_{\mathrm{eff}}) \;+\; c\cdot A_{\mathrm{eff}} .
\end{equation}

\subsection*{Color Prediction Module}
\begin{figure}
    \centering
    \includegraphics[width=1\linewidth]
    {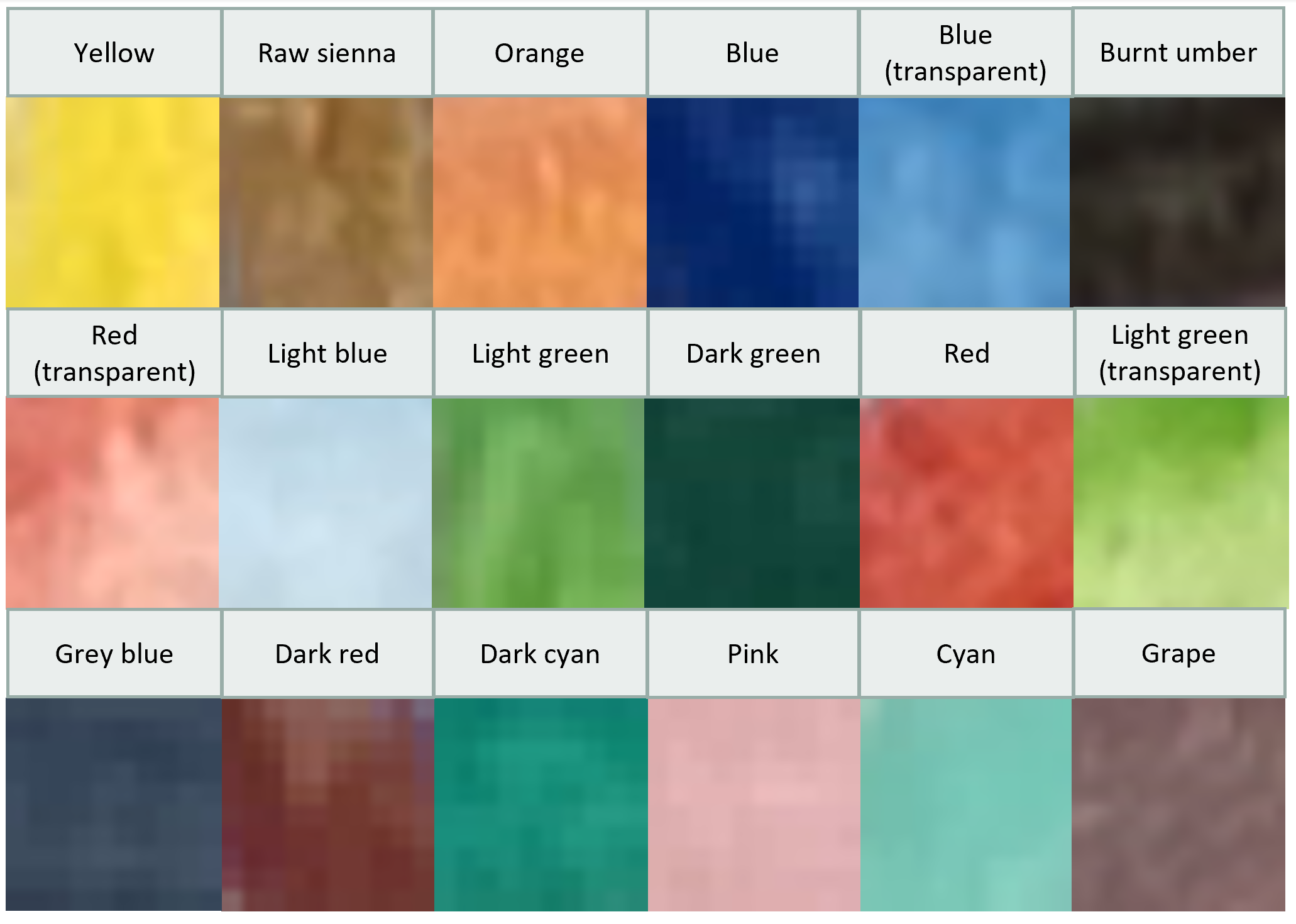}
    \caption{A subset of the color patch database.}
    \label{fig:color}
\end{figure}

The palette tray physically contains 24 slots, while only a task-specific subset of premixed colors is used in the reported experiments, depending on the target painting. The color patch database (partially shown in Fig.~\ref{fig:color}) stores multiple measured patches per color/transparency condition. In the single-stroke trajectory following pipeline, we build a difference mask \(S\) by thresholding base-target pixel differences and obtain per-pixel weights \(w_p\) from a distance transform. In the multi-step planning pipeline, we obtain $S$ as the intersection between the foreground region of the pre-rendered predicted stroke and a disk centered at the stroke’s geometric center:
\begin{equation}
S \;=\; \big\{\,p \in \Omega \;\big|\; M_{\text{pred}}(p)=1 \;\wedge\; \|p-c\|\le r \,\big\},
\end{equation}
where $M_{\text{pred}}$ is the binary mask of the pre-rendered stroke, $c$ is its geometric center, and $r$ is a chosen radius. 
The representative color for the stroke is the weighted mean over the mask,
\begin{equation}
\mu \;=\; \frac{\sum_{p\in S} w_p\, O_p}{\sum_{p\in S} w_p}\in\mathbb{R}^3,
\end{equation}

where \(O_p\) is the target image in linear RGB at pixel \(p\). We classify \(\mu\) by nearest prototype in a $25\times25$ patch database (a subset is shown in Fig.~\ref{fig:color}) using the Mahalanobis metric. Covariances \(\Sigma_i\) for each label $i$ are estimated from the corresponding patch images:
\begin{equation}
i^\star \;=\; \arg\min_{i}\; (\mu-\mu_i)^{\!\top}\,\Sigma_i^{-1}\,(\mu-\mu_i), 
\qquad C^\star=\mu_{i^\star}.
\end{equation}
Transparency is decided by a simple ratio that compares the median magnitude of the observed change to the palette-to-base contrast for the selected color:
\begin{equation}
\mathrm{ratio} \;=\; 
\frac{\operatorname{median}_{p\in S}\!\left\lVert O_p - B_p \right\rVert_{1}}
     {\operatorname{median}_{p\in S}\!\left\lVert C^\star - B_p \right\rVert_{1} + \varepsilon},
\end{equation}
where we predict transparent if $\mathrm{ratio}<\tau$,
with \(B_p\) the base image in linear RGB, \(\varepsilon>0\) for numerical safety, and threshold \(\tau\). 

\begin{figure*}[tbh!]
    \centering
    \includegraphics[width=1\linewidth]
    {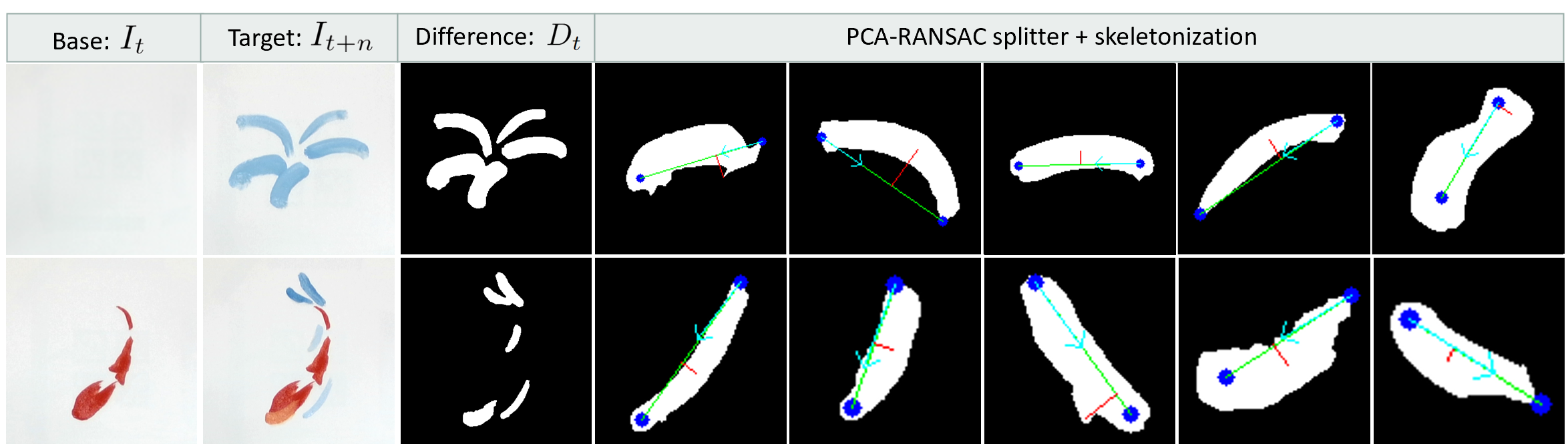}
    \caption{Example visualizations produced by the PCA-RANSAC Splitter.}
    \label{fig:splitter}
\end{figure*}

\section*{Appendix C. Planning}
\subsection*{PCA-RANSAC multi-way splitter}
The PCA--RANSAC multi-way splitter constructs \(K\) disjoint stroke masks from the difference map $D_t$ obtained from the base image $I_t$ and the target image $I_{t+n}$. We segment $D_t$ with \(k\)-means in a spatial feature space by embedding each pixel at \( [x/W,\, y/H]\) where \((x,y)\) are image coordinates and \(W\) and \(H\) are image width and height. Each resulting segment is treated as a candidate group of stroke pixels. For every segment we compute a local PCA frame; letting \(\mu \in \mathbb{R}^2\) be the segment mean and \(v_0,v_1 \in \mathbb{R}^2\) the principal directions, a pixel \((x,y)\) is projected to \((s,t)\) as
\begin{equation}
\begin{bmatrix}s\\ t\end{bmatrix}
=
\begin{bmatrix}v_0^{\!\top}\\ v_1^{\!\top}\end{bmatrix}
\!\bigl([x,y]^{\!\top}-\mu\bigr).
\end{equation}
We apply Random Sample Consensus (RANSAC) to robustly fit either a line \(t \approx p\,s + q\) or a quadratic \(t \approx p\,s^{2} + q\,s + r\) to the \((s,t)\) samples. The RANSAC inliers define a thin candidate mask. We select \(K\) non-overlapping strokes by maximizing coverage of the unexplained region with penalties on curvature and mutual overlap,
\begin{equation}
\text{score} \;=\; \text{cover} \;-\; \lambda_b\,\text{bend} \;-\; \lambda_o\,\text{overlap},
\end{equation}
and thicken each selected mask into a single-connected stroke via centerline seeding and constrained region growing, yielding \(K\) heuristic initializations for the MPPI optimization. An example of the resulting visualization from the PCA-RANSAC Splitter is shown in Fig.~\ref{fig:splitter}.

\subsection*{Stochastic Sampling with CMA-Style Adaptive Covariance}
At each MPPI optimization iteration, we assemble a candidate set by inserting the current nominal sequence $\mathbf U$ and drawing the remaining $K{-}1$ samples from a time-indexed Gaussian. Concretely, for each time index $t$ we generate standard normal noise $z_t \sim \mathcal N(0,I)$, apply the current covariance structure via a Cholesky factor $A_t$ of $\Sigma_t$, where $\Sigma_t = A_t A_t^\top$, and form perturbed controls 
\begin{equation}\,\tilde U_t = U_t + \sigma_t A_t z_t.\,\end{equation}
The full sequences are clipped to action bounds. This yields a batch of candidates centered at $\mathbf U$ but shaped by $(\sigma_t,\Sigma_t)$ at every timestep.

From all $K$ candidates, we select the top $K_e$ by terminal cost and compute temperature-weighted softmax weights with temperature $\beta$ over the elite set. Using those weights, we update the nominal sequence via an exponential moving average (EMA) toward the elite mean and perform CMA-style per-timestep adaptation: the step-size path $p_s(t)$ updates $\sigma_t$, and the covariance path $p_c(t)$ together with a rank-$\mu$ term updates $\Sigma_t$.

\section*{Appendix D. Additional Results}
\begin{figure*}[tbh!]
    \centering
    \includegraphics[width=1\linewidth]{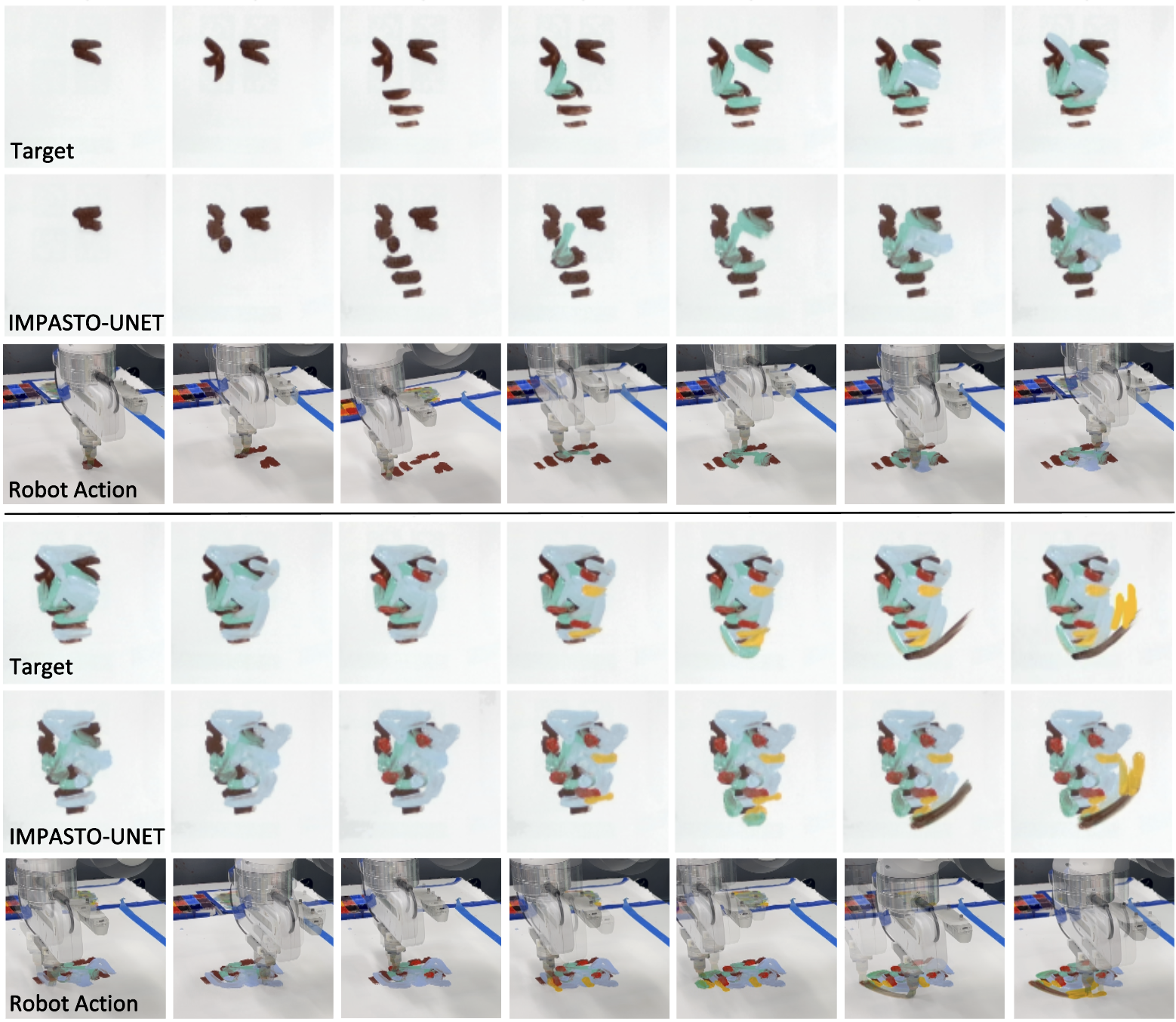}
    \caption{\algo is able to closely approximate a complex oil painting with 40 strokes (not all steps are shown).}
    \label{fig:final_demo_0}
    \vspace{-0.5cm}
\end{figure*}

\begin{figure*}[tbh!]
    \centering
    \includegraphics[width=1\linewidth]{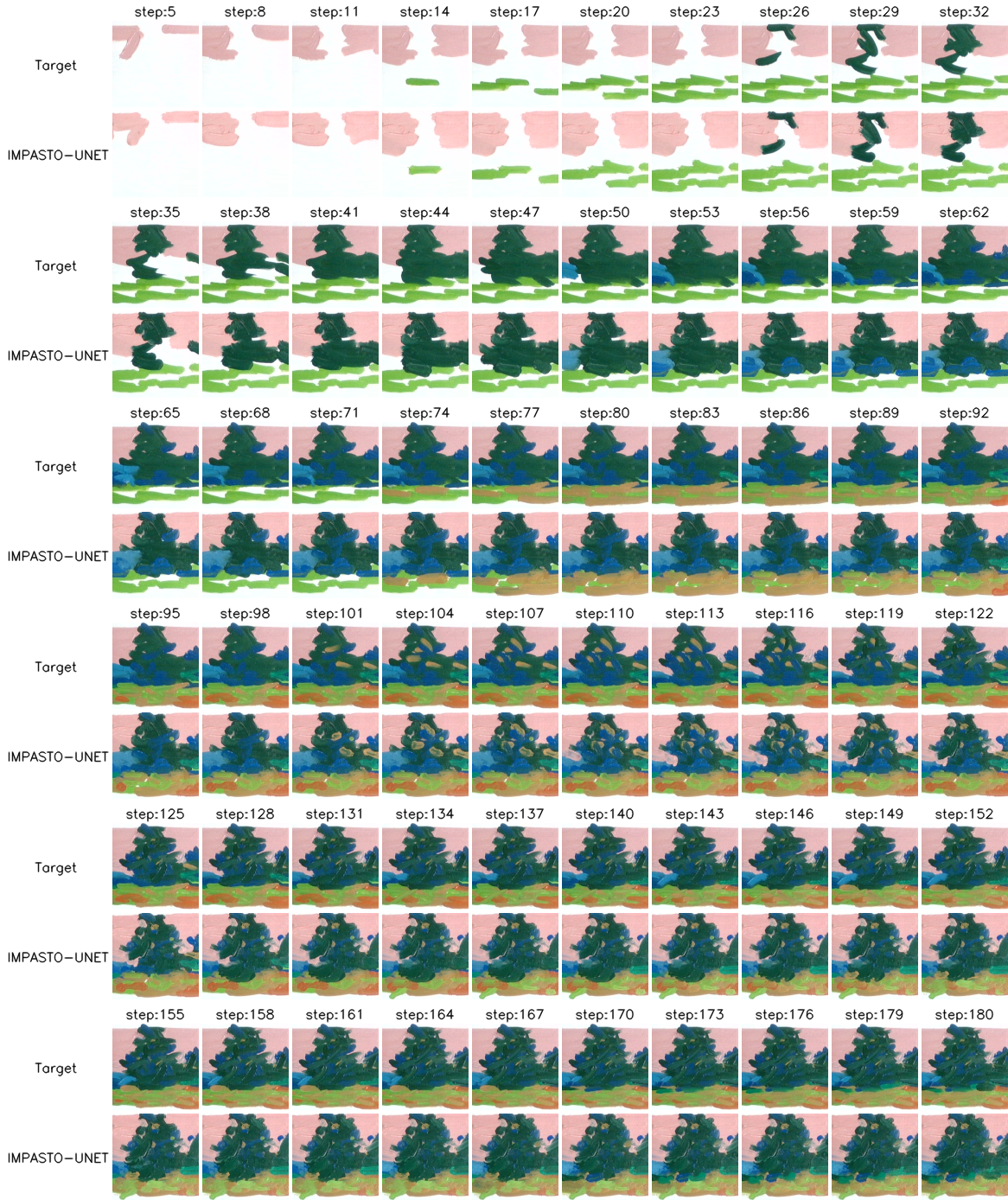}
    \caption{\algo is able to closely approximate a complex oil painting with 180 strokes (not all steps are shown).}
    \label{fig:final_demo_1}
    \vspace{-0.5cm}
\end{figure*}

\begin{figure*}[tbh!]
    \centering
    \includegraphics[width=1\linewidth]{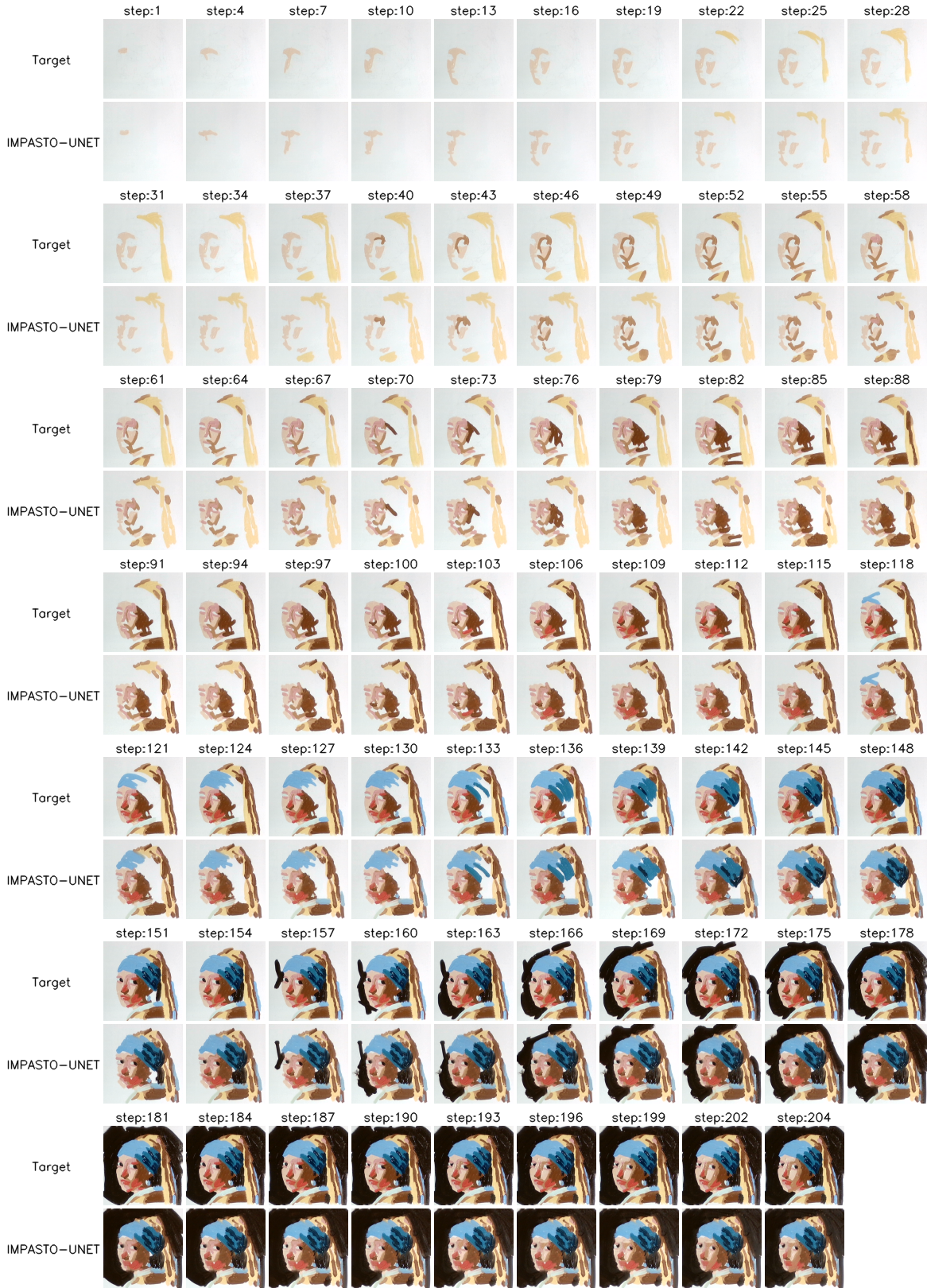}
    \caption{\algo is able to closely approximate a complex oil painting with 204 strokes (not all steps are shown).}
    \label{fig:final_demo_2}
    \vspace{-0.5cm}
\end{figure*}

For the complex paintings shown in the video, the robot follows the human artist step by step using CMA-ES to complete this artwork. Additional visualization results are shown in Fig.~\ref{fig:final_demo_0},~\ref{fig:final_demo_1}, and~\ref{fig:final_demo_2}, comparing human strokes with robot strokes. 



%% file: ref-dyn.bib
@inproceedings{zhang2024adaptigraph,
      title={AdaptiGraph: Material-Adaptive Graph-Based Neural Dynamics for Robotic Manipulation},
      author={Zhang, Kaifeng and Li, Baoyu and Hauser, Kris and Li, Yunzhu},
      booktitle={Proceedings of Robotics: Science and Systems (RSS)},
      year={2024}
    }

@inproceedings{hafner2019planet,
  title={Learning Latent Dynamics for Planning from Pixels},
  author={Hafner, Danijar and Lillicrap, Timothy and Fischer, Ian and Villegas, Ruben and Ha, David and Lee, Honglak and Davidson, James},
  booktitle={International Conference on Machine Learning},
  pages={2555--2565},
  year={2019}
}

@misc{shi2023robocook,
      title={RoboCook: Long-Horizon Elasto-Plastic Object Manipulation with Diverse Tools}, 
      author={Haochen Shi and Huazhe Xu and Samuel Clarke and Yunzhu Li and Jiajun Wu},
      year={2023},
      eprint={2306.14447},
      archivePrefix={arXiv},
      primaryClass={cs.RO}
}

@article{finn2016unsupervised,
  title={Unsupervised learning for physical interaction through video prediction},
  author={Finn, Chelsea and Goodfellow, Ian and Levine, Sergey},
  journal={arXiv preprint arXiv:1605.07157},
  year={2016}
}

@inproceedings{finn2017deep,
  title={Deep visual foresight for planning robot motion},
  author={Finn, Chelsea and Levine, Sergey},
  booktitle={2017 IEEE International Conference on Robotics and Automation (ICRA)},
  pages={2786--2793},
  year={2017},
  organization={IEEE}
}

@article{ebert2018visual,
  title={Visual foresight: Model-based deep reinforcement learning for vision-based robotic control},
  author={Ebert, Frederik and Finn, Chelsea and Dasari, Sudeep and Xie, Annie and Lee, Alex and Levine, Sergey},
  journal={arXiv preprint arXiv:1812.00568},
  year={2018}
}

@inproceedings{ebert2017self,
  title={Self-Supervised Visual Planning with Temporal Skip Connections.},
  author={Ebert, Frederik and Finn, Chelsea and Lee, Alex X and Levine, Sergey},
  booktitle={CoRL},
  year={2017}
}

@inproceedings{yen2020experience,
  title={Experience-embedded visual foresight},
  author={Yen-Chen, Lin and Bauza, Maria and Isola, Phillip},
  booktitle={CoRL},
  pages={1015--1024},
  year={2020},
  organization={PMLR}
}

@article{suh2020surprising,
  title={The surprising effectiveness of linear models for visual foresight in object pile manipulation},
  author={Suh, HJ and Tedrake, Russ},
  journal={arXiv preprint arXiv:2002.09093},
  year={2020}
}

@article{watter2015embed,
  title={Embed to control: A locally linear latent dynamics model for control from raw images},
  author={Watter, Manuel and Springenberg, Jost Tobias and Boedecker, Joschka and Riedmiller, Martin},
  journal={arXiv preprint arXiv:1506.07365},
  year={2015}
}

@article{agrawal2016learning,
  title={Learning to poke by poking: Experiential learning of intuitive physics},
  author={Agrawal, Pulkit and Nair, Ashvin and Abbeel, Pieter and Malik, Jitendra and Levine, Sergey},
  journal={arXiv preprint arXiv:1606.07419},
  year={2016}
}

@article{hafner2019dream,
  title={Dream to control: Learning behaviors by latent imagination},
  author={Hafner, Danijar and Lillicrap, Timothy and Ba, Jimmy and Norouzi, Mohammad},
  journal={arXiv preprint arXiv:1912.01603},
  year={2019}
}

@article{schrittwieser2020mastering,
  title={Mastering atari, go, chess and shogi by planning with a learned model},
  author={Schrittwieser, Julian and Antonoglou, Ioannis and Hubert, Thomas and Simonyan, Karen and Sifre, Laurent and Schmitt, Simon and Guez, Arthur and Lockhart, Edward and Hassabis, Demis and Graepel, Thore and others},
  journal={Nature},
  volume={588},
  number={7839},
  pages={604--609},
  year={2020},
  publisher={Nature Publishing Group}
}

@inproceedings{wu2023daydreamer,
  title={Daydreamer: World models for physical robot learning},
  author={Wu, Philipp and Escontrela, Alejandro and Hafner, Danijar and Abbeel, Pieter and Goldberg, Ken},
  booktitle={CoRL},
  year={2023},
}

@article{kulkarni2019unsupervised,
  title={Unsupervised learning of object keypoints for perception and control},
  author={Kulkarni, Tejas D and Gupta, Ankush and Ionescu, Catalin and Borgeaud, Sebastian and Reynolds, Malcolm and Zisserman, Andrew and Mnih, Volodymyr},
  journal={Advances in neural information processing systems},
  volume={32},
  pages={10724--10734},
  year={2019}
}

@article{li2020causal,
    title={Causal discovery in physical systems from videos},
    author={Li, Yunzhu and Torralba, Antonio and Anandkumar, Anima and Fox, Dieter and Garg, Animesh},
    journal={Advances in Neural Information Processing Systems},
    volume={33},
    year={2020}
}

@article{shi2022robocraft,
  title={RoboCraft: Learning to See, Simulate, and Shape Elasto-Plastic Objects with Graph Networks},
  author={Shi, Haochen and Xu, Huazhe and Huang, Zhiao and Li, Yunzhu and Wu, Jiajun},
  journal={arXiv preprint arXiv:2205.02909},
  year={2022}
}

@inproceedings{huang2022medor,
	title={Mesh-based Dynamics Model with Occlusion Reasoning for Cloth Manipulation},
	author={Huang, Zixuan and Lin, Xingyu and Held,David},
	booktitle={RSS},
	year={2022}
}

@misc{han2024model,
      title={Model Predictive Control for Aggressive Driving Over Uneven Terrain}, 
      author={Tyler Han and Alex Liu and Anqi Li and Alex Spitzer and Guanya Shi and Byron Boots},
      year={2024},
      eprint={2311.12284},
      archivePrefix={arXiv},
      primaryClass={cs.RO}
}

@misc{sacks2023deep,
      title={Deep Model Predictive Optimization}, 
      author={Jacob Sacks and Rwik Rana and Kevin Huang and Alex Spitzer and Guanya Shi and Byron Boots},
      year={2023},
      eprint={2310.04590},
      archivePrefix={arXiv},
      primaryClass={cs.RO}
}

@inproceedings{nagabandi2020deep,
  title={Deep dynamics models for learning dexterous manipulation},
  author={Nagabandi, Anusha and Konolige, Kurt and Levine, Sergey and Kumar, Vikash},
  booktitle={CoRL},
  year={2020},
}

@article{manuelli2020keypoints,
  title={Keypoints into the Future: Self-Supervised Correspondence in Model-Based Reinforcement Learning},
  author={Manuelli, Lucas and Li, Yunzhu and Florence, Pete and Tedrake, Russ},
  journal={arXiv preprint arXiv:2009.05085},
  year={2020}
}

@book{rubinstein2013cross,
  title={The cross-entropy method: a unified approach to combinatorial optimization, Monte-Carlo simulation and machine learning},
  author={Rubinstein, Reuven Y and Kroese, Dirk P},
  year={2013},
  publisher={Springer Science \& Business Media}
}

@article{williams2017model,
  title={Model predictive path integral control: From theory to parallel computation},
  author={Williams, Grady and Aldrich, Andrew and Theodorou, Evangelos A},
  journal={Journal of Guidance, Control, and Dynamics},
  pages={344--357},
  year={2017},
  publisher={American Institute of Aeronautics and Astronautics}
}

@inproceedings{li2019propagation,
  title={Propagation networks for model-based control under partial observation},
  author={Li, Yunzhu and Wu, Jiajun and Zhu, Jun-Yan and Tenenbaum, Joshua B and Torralba, Antonio and Tedrake, Russ},
  booktitle={ICRA},
  pages={1205--1211},
  year={2019},
  organization={IEEE}
}

@article{li2018learning,
  title={Learning particle dynamics for manipulating rigid bodies, deformable objects, and fluids},
  author={Li, Yunzhu and Wu, Jiajun and Tedrake, Russ and Tenenbaum, Joshua B and Torralba, Antonio},
  journal={arXiv preprint arXiv:1810.01566},
  year={2018}
}

@inproceedings{wang2023dynamic,
  title={Dynamic-Resolution Model Learning for Object Pile Manipulation},
  author={Wang, Yixuan and Li, Yunzhu and Driggs-Campbell, Katherine and Fei-Fei, Li and Wu, Jiajun},
  booktitle={RSS},
  year={2023}
}


%% file: ref.bib
@article{cohen1995further,
  title={The further exploits of AARON, painter},
  author={Cohen, Harold},
  journal={Stanford Humanities Review},
  volume={4},
  number={2},
  pages={141--158},
  year={1995}
}

@inproceedings{deussen2012feedback,
  title={Feedback-guided stroke placement for a painting machine},
  author={Deussen, Oliver and Lindemeier, Thomas and Pirk, S{\"o}ren and Tautzenberger, Mark},
  booktitle={Proceedings of the Eighth Annual Symposium on Computational Aesthetics in Graphics, Visualization, and Imaging},
  pages={25--33},
  year={2012}
}

@article{lindemeier2013image,
  title={Image stylization with a painting machine using semantic hints},
  author={Lindemeier, Thomas and Pirk, S{\"o}ren and Deussen, Oliver},
  journal={Computers \& Graphics},
  volume={37},
  number={5},
  pages={293--301},
  year={2013},
  publisher={Elsevier}
}

@inproceedings{schaldenbrand2023frida,
  title={FRIDA: A Collaborative Robot Painter with a Differentiable, Real2Sim2Real Planning Environment},
  author={Schaldenbrand, Peter and McCann, James and Oh, Jean},
  booktitle={2023 IEEE International Conference on Robotics and Automation (ICRA)},
  pages={11712--11718},
  year={2023},
  organization={IEEE}
}

@inproceedings{schaldenbrand2024cofrida,
  title={Cofrida: Self-supervised fine-tuning for human-robot co-painting},
  author={Schaldenbrand, Peter and Parmar, Gaurav and Zhu, Jun-Yan and McCann, James and Oh, Jean},
  booktitle={ICRA},
  pages={2296--2302},
  year={2024},
  organization={IEEE}
}

@article{chen2025spline,
  title={Spline-FRIDA: Towards Diverse, Humanlike Robot Painting Styles with a Sample-Efficient, Differentiable Brush Stroke Model},
  author={Chen, Lawrence and Schaldenbrand, Peter and Shankar, Tanmay and Coleman, Lia and Oh, Jean},
  journal={IEEE RAL},
  year={2025},
  publisher={IEEE}
}

@inproceedings{scalera2024history,
  title={History of drawing robots},
  author={Scalera, Lorenzo and Gasparetto, Alessandro and Seriani, Stefano and Gallina, Paolo},
  booktitle={International Symposium on History of Machines and Mechanisms},
  pages={3--17},
  year={2024},
  organization={Springer}
}

@article{berrocal5396199reinforcement,
  title={Reinforcement Learning for Autonomous Surface Painting with a Robotic Arm},
  author={Berrocal, Enaitz and Merino, Ibon and Monta{\~n}o, Andr{\'e}s Felipe and Sierra, Basilio},
  journal={Available at SSRN 5396199}
}

@inproceedings{tiboni2023paintnet,
  title={PaintNet: Unstructured multi-path learning from 3D point clouds for robotic spray painting},
  author={Tiboni, Gabriele and Camoriano, Raffaello and Tommasi, Tatiana},
  booktitle={2023 IROS},
  pages={3857--3864},
  year={2023},
  organization={IEEE}
}

@inproceedings{schaldenbrand2021content,
  title={Content masked loss: Human-like brush stroke planning in a reinforcement learning painting agent},
  author={Schaldenbrand, Peter and Oh, Jean},
  booktitle={AAAI},
  year={2021}
}

@inproceedings{lee2022scratch,
  title={From scratch to sketch: Deep decoupled hierarchical reinforcement learning for robotic sketching agent},
  author={Lee, Ganghun and Kim, Minji and Lee, Minsu and Zhang, Byoung-Tak},
  booktitle={2022 International Conference on Robotics and Automation (ICRA)},
  pages={5553--5559},
  year={2022},
  organization={IEEE}
}

@article{jia2023sim,
  title={Sim-to-Real Brush Manipulation using Behavior Cloning and Reinforcement Learning},
  author={Jia, Biao and Manocha, Dinesh},
  journal={arXiv preprint arXiv:2309.08457},
  year={2023}
}

@inproceedings{park2022robot,
  title={Robot learning to paint from demonstrations},
  author={Park, Younghyo and Jeon, Seunghun and Lee, Taeyoon},
  booktitle={2022 IEEE/RSJ international conference on intelligent robots and systems (IROS)},
  pages={3053--3060},
  year={2022},
  organization={IEEE}
}

@article{guo2022shadowpainter,
  title={ShadowPainter: Active learning enabled robotic painting through visual measurement and reproduction of the artistic creation process},
  author={Guo, Chao and Bai, Tianxiang and Wang, Xiao and Zhang, Xiangyu and Lu, Yue and Dai, Xingyuan and Wang, Fei-Yue},
  journal={Journal of Intelligent \& Robotic Systems},
  volume={105},
  number={3},
  pages={61},
  year={2022},
  publisher={Springer}
}

@inproceedings{liu2021paint,
  title={Paint transformer: Feed forward neural painting with stroke prediction},
  author={Liu, Songhua and Lin, Tianwei and He, Dongliang and Li, Fu and Deng, Ruifeng and Li, Xin and Ding, Errui and Wang, Hao},
  booktitle={ICCV},
  pages={6598--6607},
  year={2021}
}

@article{karimov2023robot,
  title={A robot for artistic painting in authentic colors},
  author={Karimov, Artur and Kopets, Ekaterina and Leonov, Sergey and Scalera, Lorenzo and Butusov, Denis},
  journal={Journal of Intelligent \& Robotic Systems},
  volume={107},
  number={3},
  pages={34},
  year={2023},
  publisher={Springer}
}

@article{li2020differentiable,
  title={Differentiable vector graphics rasterization for editing and learning},
  author={Li, Tzu-Mao and Luk{\'a}{\v{c}}, Michal and Gharbi, Micha{\"e}l and Ragan-Kelley, Jonathan},
  journal={ACM Transactions on Graphics (TOG)},
  volume={39},
  number={6},
  pages={1--15},
  year={2020},
  publisher={ACM New York, NY, USA}
}

@article{gulzow2020recent,
  title={Recent developments regarding painting robots for research in automatic painting, artificial creativity, and machine learning},
  author={G{\"u}lzow, J{\"o}rg Marvin and Paetzold, Patrick and Deussen, Oliver},
  journal={Applied Sciences},
  volume={10},
  number={10},
  pages={3396},
  year={2020},
  publisher={MDPI}
}

@inproceedings{wu2018brush,
  title={Brush stroke synthesis with a generative adversarial network driven by physically based simulation},
  author={Wu, Rundong and Chen, Zhili and Wang, Zhaowen and Yang, Jimei and Marschner, Steve},
  booktitle={Proceedings of the joint symposium on computational aesthetics and sketch-based interfaces and modeling and non-photorealistic animation and rendering},
  pages={1--10},
  year={2018}
}

@article{guo2024b,
  title={B-BSMG: B{\'e}zier Brush Stroke Model-Based Generator for Robotic Chinese Calligraphy},
  author={Guo, Dongmei and Yan, Guang},
  journal={International Journal of Computational Intelligence Systems},
  volume={17},
  number={1},
  pages={104},
  year={2024},
  publisher={Springer}
}

@inproceedings{aguilar2008robotic,
  title={A robotic system for interpreting images into painted artwork},
  author={Aguilar, Carlos and Lipson, Hod},
  booktitle={International conference on generative art},
  volume={11},
  year={2008}
}

@inproceedings{shenbab,
  title={BaB-ND: Long-Horizon Motion Planning with Branch-and-Bound and Neural Dynamics},
  author={Shen, Keyi and Yu, Jiangwei and Barreiros, Jose and Zhang, Huan and Li, Yunzhu},
  booktitle={The Thirteenth ICLR}
}

@inproceedings{wang2020robot,
  title={Robot calligraphy using pseudospectral optimal control in conjunction with a novel dynamic brush model},
  author={Wang, Sen and Chen, Jiaqi and Deng, Xuanliang and Hutchinson, Seth and Dellaert, Frank},
  booktitle={IROS},
  pages={6696--6703},
  year={2020},
  organization={IEEE}
}

@inproceedings{ronneberger2015u,
  title={U-net: Convolutional networks for biomedical image segmentation},
  author={Ronneberger, Olaf and Fischer, Philipp and Brox, Thomas},
  booktitle={International Conference on Medical image computing and computer-assisted intervention},
  pages={234--241},
  year={2015},
  organization={Springer}
}

@inproceedings{zhang2018unreasonable,
  title={The unreasonable effectiveness of deep features as a perceptual metric},
  author={Zhang, Richard and Isola, Phillip and Efros, Alexei A and Shechtman, Eli and Wang, Oliver},
  booktitle={CVPR},
  pages={586--595},
  year={2018}
}

@article{de2000mahalanobis,
  title={The mahalanobis distance},
  author={De Maesschalck, Roy and Jouan-Rimbaud, Delphine and Massart, D{\'e}sir{\'e} L},
  journal={Chemometrics and intelligent laboratory systems},
  volume={50},
  number={1},
  pages={1--18},
  year={2000},
  publisher={Elsevier}
}
